%% file: bare_jrnl_new_sample4.tex
\documentclass[lettersize,journal]{IEEEtran}
\usepackage{amsmath,amsfonts}
\usepackage{algorithmic}
\usepackage{algorithm}
\usepackage{array}
\usepackage{textcomp}
\usepackage{stfloats}
\usepackage{url}
\usepackage{verbatim}
\usepackage{graphicx}
\usepackage{subcaption}
\usepackage{graphicx}    
\usepackage{subcaption}  
\usepackage{float}       
\usepackage{float} 
\usepackage{booktabs}
\usepackage{caption}
\usepackage{cite}
\usepackage{multirow}
\usepackage{xcolor}
\def\netName{TaPD}

\begin{document}

\title{TaPD: Temporal-adaptive Progressive Distillation for Observation-Adaptive Trajectory Forecasting in Autonomous Driving}

\author{Mingyu Fan, ~\IEEEmembership{Member,~IEEE,} Yi Liu,  Hao Zhou$^*$, Deheng Qian, Mohammad Haziq Khan, Matthias Raetsch
\thanks{Mingyu Fan, Yi Liu are with the College of Information and Intelligent Science, Donghua University, Shanghai 201620 China.}
\thanks{Hao Zhou is with the School of Computing and Information Technology,
Great Bay Institute for Advanced Study/Great Bay University, Dongguan, China, and also with the Tsinghua Shenzhen International Graduate School, Tsinghua University, Shenzhen, China.}
\thanks{Deheng Qian is with the Chongqing Chang'an Technology Co., Ltd, Chongqing 401120‌ China.}
\thanks{Mohammad Haziq Khan and Matthias Rätsch are with the ViSiR, Reutlingen University, Alteburgstraße 150, 72762, Reutlingen, Germany.}
\thanks{Corresponding author: Hao Zhou (e-mail: hao.zhou@hrbeu.edu.cn).}}


\newcommand{\zh}[1]{#1}
\maketitle

\begin{abstract}
Trajectory prediction is essential for autonomous driving, enabling vehicles to anticipate the motion of surrounding agents to support safe planning. However, most existing predictors assume fixed-length histories and suffer substantial performance degradation when observations are variable or extremely short in real-world settings (e.g., due to occlusion or a limited sensing range). We propose TaPD (Temporal-adaptive Progressive Distillation), a unified plug-and-play framework for observation-adaptive trajectory forecasting under variable history lengths. TaPD comprises two cooperative modules: an Observation-Adaptive Forecaster (OAF) for future prediction and a Temporal Backfilling Module (TBM) for explicit reconstruction of the past. OAF is built on progressive knowledge distillation (PKD), which transfers motion pattern knowledge from long-horizon “teachers” to short-horizon “students” via hierarchical feature regression, enabling short observations to recover richer motion context. We further introduce a cosine-annealed distillation weighting scheme to balance forecasting supervision and feature alignment, improving optimization stability and cross-length consistency. For extremely short histories where implicit alignment is insufficient, TBM backfills missing historical segments conditioned on scene evolution, producing context-rich trajectories that \zh{strengthen PKD and thereby improve OAF}. We employ a decoupled pretrain–reconstruct–finetune protocol to preserve real-motion priors while adapting to backfilled inputs. Extensive experiments on Argoverse 1 and Argoverse 2 show that TaPD consistently outperforms strong baselines across all observation lengths, delivers especially large gains under very short inputs, and improves other predictors (e.g., HiVT) in a plug-and-play manner. \zh{Code will be available at \url{https://github.com/zhouhao94/TaPD}}.
\end{abstract}

\begin{IEEEkeywords}
Trajectory prediction, autonomous driving, observation-adaptive learning.
\end{IEEEkeywords}

\section{Introduction}

\IEEEPARstart{T}{rajectory} prediction plays a vital role in autonomous driving, allowing an ego vehicle to anticipate the future motions of dynamic agents (e.g., vehicles, pedestrians, cyclists) and thereby support safe and socially compliant planning \cite{yurtsever2020survey, Katrakazas_Quddus_Chen_Deka_2015}. Recent learning-based predictors \cite{mtr, qcnet, smartrefine, wayformer, prophnet} have achieved strong performance by exploiting high-definition (HD) maps and attention-based interaction modeling. Nevertheless, these methods are predominantly trained and benchmarked with \emph{fixed-length} observation windows, which diverges from real deployments where the available history is inherently \emph{variable}. In practice, agents may be intermittently occluded, partially sensed, or appear abruptly at the boundary of the perception range (e.g., a pedestrian emerging from behind a parked vehicle). Under such conditions, the predictor must operate on severely truncated trajectories (sometimes only a handful of steps), where the loss of motion context can cause substantial accuracy drops and increase downstream safety risks.

A straightforward solution is \emph{Isolated Training (IT)}, i.e., training one model per observation length. While effective in principle, IT is inefficient and deployment-unfriendly due to duplicated parameters, repeated training, and additional maintenance overhead. More efficient approaches based on cross-length parameter sharing (e.g., FLN \cite{FLN}) partially alleviate the computational burden, yet they often remain brittle for extremely short histories. The core difficulty is not merely architectural: \emph{short histories suffer from an intrinsic information deficit}. Pure feature-level alignment across lengths cannot reliably recover trajectory-specific prior states (e.g., heading, velocity trend, pre-occlusion maneuver intention) that are unobserved, leading to a persistent distribution gap between short- and long-history representations.

To address these challenges, we propose \textbf{TaPD} (\textbf{T}emporal-\textbf{a}daptive \textbf{P}rogressive \textbf{D}istillation), a unified and plug-and-play framework for \emph{observation-adaptive trajectory forecasting}. TaPD couples two cooperative modules: an \textbf{Observation-Adaptive Forecaster (OAF)} for future prediction and a \textbf{Temporal Backfilling Module (TBM)} for explicit history completion. OAF is designed to handle arbitrary observation lengths within a single network via \emph{cross-length parameter sharing}, avoiding the redundancy of IT while preserving representation consistency between training and inference. On top of this shared backbone, OAF introduces \emph{progressive knowledge distillation (PKD)}: long-history features serve as teachers that guide short-history features through hierarchical regression, enabling short inputs to inherit richer motion-pattern knowledge and implicitly compensate for missing context. To prevent early-stage optimization instability when motion representations are still immature, we further employ a \emph{cosine-annealed} weighting schedule for the distillation objective, emphasizing trajectory supervision at the beginning and gradually strengthening feature alignment as training proceeds.

While PKD improves cross-length robustness, extremely short observations may still lack sufficient state information for implicit alignment alone. TBM therefore complements OAF by \emph{explicit temporal backfilling}: conditioned on the evolving scene context, it reconstructs the missing historical segments and converts truncated trajectories into standardized full-length inputs. This explicit completion supplies trajectory-specific priors that feature matching cannot reliably infer, and in turn enables OAF to exploit PKD more effectively on context-rich histories.

To fully realize this synergy without corrupting learned motion priors, we adopt a decoupled \emph{pretrain--reconstruct--finetune} protocol: (i) we pretrain OAF on real trajectories to learn forecasting and cross-length adaptation; (ii) we train TBM independently to produce high-fidelity backfilled histories; and (iii) we freeze TBM and finetune OAF to adapt to completed inputs while retaining authentic motion regularities. With this design, TaPD can be integrated into existing trajectory prediction pipelines with minimal changes, providing robust performance under arbitrary observation lengths.

The main contributions of this work are summarized as follows:
\begin{itemize}
\item We propose TaPD, a unified dual-module framework for \emph{observation-adaptive} trajectory forecasting that remains robust under arbitrary and extremely short observation histories.
\item We design an OAF module with \emph{cross-length parameter sharing} and \emph{PKD}, enabling efficient knowledge transfer from long to short histories without training separate models, and also a \emph{cosine-annealed distillation schedule} to stabilize training and progressively enhance cross-length feature consistency.
\item We propose a TBM module that explicitly reconstructs missing historical segments, provides trajectory-specific priors, and synergistically strengthens OAF under extreme truncation.
\item We develop a decoupled \emph{pretrain--reconstruct--finetune} training protocol and demonstrate, through extensive experiments on Argoverse~1 and Argoverse~2, that TaPD consistently outperforms strong baselines across all observation lengths and can improve other predictors in a plug-and-play manner.
\end{itemize}

\section{Related Works}

Trajectory prediction (motion forecasting) is a cornerstone of autonomous driving stacks, providing multi-hypothesis futures for risk-aware planning and safe interaction in dynamic traffic \cite{yurtsever2020survey,Katrakazas_Quddus_Chen_Deka_2015}. With the availability of large-scale benchmarks (e.g., Argoverse~1/2\cite{argoverse,argoverse2} and WOMD\cite{WOMD}), learning-based predictors have rapidly evolved in scene representation, interaction modeling, and generative uncertainty modeling. This section reviews related efforts from three perspectives: (i) representations and architectures, (ii) training paradigms for multimodality, efficiency, and robustness, and (iii) forecasting under partial observability and variable observation lengths.

\subsection{Representations and Architectures for Motion Forecasting}
\textbf{Raster-based predictors.}
Early deep forecasting pipelines often rasterize agent states and HD maps into images and apply CNN backbones to regress future trajectories or trajectory sets \cite{multipath,covernet,home}. Rasterization simplifies heterogeneous fusion but may obscure fine-grained topology and lane connectivity, motivating structured representations. To better preserve road structure, vectorized polyline encoders and lane-graph reasoning have become mainstream. VectorNet \cite{vectornet} demonstrates the effectiveness of polyline-level encoding for jointly modeling agent dynamics and map geometry. Lane graph modeling further strengthens topological priors by propagating context along lane connectivity and candidate centerlines \cite{LaneGCN,lanercnn,paga}. In broader trajectory prediction settings, graph formulations also capture social interactions and global context effectively \cite{zhouStatic2023,zhouCSIR2023,TKDEtdagcn,TKDEasgtn,TKDEAisfuser}. Recently, transformers \cite{transformer} have become the dominant paradigm for modeling complex agent--agent and agent--map interactions. Representative works include HiVT \cite{hivt} and MTR \cite{mtr}, which leverage hierarchical/vector attention for multi-agent reasoning, and Wayformer \cite{wayformer}, which highlights the scalability of efficient attention designs. Query/goal-driven formulations further improve multimodal forecasting by decoupling \emph{where} to go and \emph{how} to move, such as TNT/DenseTNT \cite{tnt,densetnt}, anchor/goal proposals \cite{ganet,prophnet}, and query-centric decoding \cite{qcnet}. Meanwhile, efficiency-oriented modeling has attracted increasing attention. For example, Trajectory Mamba \cite{huang2025trajectory} explores selective state-space modeling to reduce the quadratic complexity of attention for forecasting.

Recent works indicate two notable trends. First, decoder-only autoregressive forecasting (DONUT) unifies history encoding and future unrolling in a single model, improving iterative consistency for trajectory generation \cite{Knoche_2025_ICCV_DONUT}. Second, incorporating efficient SSM modules into interaction modeling is further explored by FINet \cite{Li_2025_ICCV_FINet}, which injects potential futures into scene encoding and employs Mamba-style temporal refinement.

\subsection{Learning Paradigms: Pretraining, Diffusion, and Robustness}

To mitigate dataset scarcity and enhance transferability, masked reconstruction and self-supervised objectives have been introduced for motion forecasting \cite{Forecast-mae,Traj-mae,ssl,sept}. More recently, SmartPretrain \cite{Zhou_2025_ICLR_SmartPretrain} proposes a model-agnostic and dataset-agnostic SSL framework that combines contrastive and reconstructive learning, improving cross-dataset generalization. Instead of producing final futures in a single forward pass, refinement-based pipelines first generate coarse hypotheses and then refine them using additional context or trajectory-centric attention, improving accuracy and efficiency \cite{smartrefine,R-Pred,hpnet}.

Diffusion-based forecasting has been adopted to better represent multimodal futures and controllability \cite{motiondiffuser,bcdiff}. Beyond vanilla diffusion modeling, long-tail scenarios have become a key challenge for real-world autonomy. GALTraj \cite{Park_2025_ICCV_GALTraj} introduces generative active learning with controllable diffusion to identify and augment rare behaviors during training, improving both tail and head performance. In a related direction, language-conditioned diffusion simulation (LangTraj) enables controllable scenario generation for evaluation and counterfactual testing \cite{Chang_2025_ICCV_LANGTRAJ}.

Robustness under domain shift is critical for deployment. T4P \cite{Park_2024_CVPR_T4P} explores test-time training for trajectory prediction via masked autoencoding and actor-specific token memory, demonstrating gains under cross-dataset distribution shifts. Such test-time adaptation complements training-time robustness and is increasingly relevant for real-world operation.

\subsection{Forecasting with Partial Observability and Variable Observation Lengths}
Real driving scenes rarely provide a clean and fixed-length history. Occlusion, sensor range limits, missing frames, or late entry of agents lead to truncated and irregular observations, which can significantly degrade predictors trained under fixed windows.

Some works focus on forecasting from extremely short observations, such as human motion prediction from momentary cues \cite{human} and distillation-based strategies that analyze how many observations are sufficient \cite{howmany}. While effective for specific short-history settings, these approaches do not necessarily provide a unified mechanism to handle \emph{arbitrary} observation lengths in a single model. Another line jointly performs imputation and prediction to compensate for missing segments. Uncovering the missing pattern \cite{uncovering} proposes a unified framework for trajectory imputation and forecasting, and Scene Informer \cite{sceneinformer} performs anchor-based occlusion inference in partially observable environments. Target-driven self-distillation \cite{target} further leverages auxiliary targets to guide learning under partial observations. These methods emphasize recovering missing measurements, yet cross-length generalization is often not treated as a first-class objective.

To support diverse observation lengths efficiently, length-robust learning has attracted growing attention. LaKD \cite{lakd} proposes length-agnostic distillation to transfer knowledge across observation regimes. Contrastive learning is adopted to extract length-invariant representations \cite{adapting}. FLN \cite{FLN} introduces a parameter-sharing paradigm to adapt to length shift. In parallel, dynamically adjusting output horizons is explored by FlexiSteps \cite{adaptive}. Despite these advances, extremely short histories remain challenging due to an intrinsic information deficit: purely implicit feature alignment may be insufficient to recover trajectory-specific prior states without explicitly reconstructing missing context.

\subsection{Summary}
Recent surveys summarize the fast-evolving landscape and highlight persistent gaps between benchmark settings and real-world deployment \cite{Madjid_2025_TrajSurvey}. In particular, achieving \emph{both} (i) efficient \emph{unified} modeling across \emph{arbitrary} observation lengths and (ii) strong performance under \emph{extremely short} histories remains nontrivial. This motivates our work, which targets observation-adaptive forecasting by jointly addressing cross-length knowledge transfer and the information deficit of ultra-short trajectories.

\section{TaPD: a plug-and-play framework}

This section introduces \netName{}, a unified framework for observation-adaptive motion forecasting under variable-length histories. \netName{} comprises two decoupled yet complementary components: the Observation-Adaptive Forecaster (OAF) and the Temporal Backfilling Module (TBM). OAF predicts future trajectories and learns length-robust representations via intra-module parameter sharing and progressive knowledge distillation. TBM mitigates the intrinsic information deficit of truncated histories by explicitly reconstructing unobserved past segments, converting short temporal inputs into context-rich histories. To leverage their synergy without causing mutual interference, we adopt a staged training protocol: (i) pretrain OAF to learn reliable forecasting priors and cross-length generalization, (ii) train TBM independently for high-fidelity backfilling, and (iii) freeze TBM and finetune OAF on TBM-completed histories.

\subsection{Problem Formulation}
Motion forecasting aims to predict the future trajectories of dynamic agents in driving scenes, providing multi-step motion hypotheses for safe and informed planning. Following vectorized forecasting pipelines \cite{vectornet}, we represent both the HD map and agent trajectories as structured polylines.

\textbf{Map representation.} The HD map is encoded as $\mathbf{M}\in\mathbb{R}^{P\times S_m\times C_m}$, where $P$ is the number of map polylines, each polyline is discretized into $S_m$ segments, and $C_m$ denotes the feature dimension of each segment.

\zh{\textbf{Agent history with variable observation length.} Let $\Delta T$ denote the sampling interval, and $H$ be the number of intervals of duration $\Delta T$.
The \emph{standard} full-length observation time span is $T_{\mathrm{obs}} = H\Delta T$.}

\zh{For a variable number of observed intervals $\tau\in\{1,2,\ldots,H\}$, the observed historical trajectories are}
\begin{equation}
\mathbf{X}^{\tau}\in\mathbb{R}^{N\times T_{\tau}\times C_a},
\end{equation}
where $N$ is the number of agents in the scene, $T_{\tau}=\tau\cdot\Delta T$ denotes the duration of the observed history, and $C_a$ is the motion-state dimension (e.g., position, velocity, heading). In practice, $\mathbf{X}^{\tau}$ is obtained by extracting the last $T_{\tau}$ timesteps of the full observation history $\mathbf{X}^{H}$ (i.e., the segment corresponding to time steps $[(H-\tau)\Delta T + 1 : T_{\mathrm{obs}}]$, a.k.a. $[T_{\mathrm{obs}}-T_{\tau} + 1 : T_{\mathrm{obs}}]$), which is equivalent to applying a sliding window with a fixed \zh{observation end time.}

\textbf{Prediction target.}
Given the observed history, the forecasting target is the future trajectory tensor from timesteps $[T_{obs}+1:T_{obs}+T_f]$
\begin{equation}
\mathbf{Y}\in\mathbb{R}^{N_{\mathrm{aoi}}\times T_f\times 2},
\end{equation}
where $N_{\mathrm{aoi}}$ denotes the number of agents of interest (AOIs), $T_f$ is the prediction horizon in time steps, and the last dimension corresponds to 2D coordinates.

\textbf{Temporal backfilling and forecasting.}
In our framework, the TBM aims to reconstruct missing historical segments and produce a complete full-length history
\begin{equation}
\hat{\mathbf{X}}^{H}\in\mathbb{R}^{N\times T_{obs} \times C_a},
\end{equation}
from any truncated input $\mathbf{X}^{\tau}$ with $\tau<H$. The completed history $\hat{\mathbf{X}}^{H}$ is then used as an enriched input to the OAF for forward forecasting, which generates the final future predictions $\mathbf{Y}$ for AOIs.

\input{figure/fig_main}

\subsection{Observation-Adaptive Forecaster (OAF)}

The OAF module is designed to (i) accommodate variable-length histories within a single predictor and (ii) learn reliable motion priors from real trajectories. OAF achieves this through two key mechanisms: \emph{intra-module parameter sharing (PS)} and \emph{progressive knowledge distillation (PKD)}. Architecturally, OAF consists of an encoder $\Phi_{\rm E}^{\rm OAF}$ and a decoder $\Phi_{\rm D}^{\rm OAF}$. The core network parameters are shared across all observation lengths to avoid training separate models, while length-specific LayerNorm parameters are maintained to absorb distribution shifts induced by different history lengths. This design provides an effective trade-off between computational efficiency and cross-length generalization.

\paragraph{Encoder--decoder forecasting with cross-length parameter sharing.}
Given an observed history $\mathbf{X}^{\tau}$ and the HD map $\mathbf{M}$, the encoder extracts fused scene-motion features. Specifically, feature fusion, attention, and dynamic encoding layers share parameters $\Theta_{\rm E\text{-}shared}^{\rm OAF}$ for all $\tau\in\{1,\ldots,H\}$, whereas LayerNorm parameters are length-dependent, denoted by $\Theta_{\rm E\text{-}LN}^{\rm OAF,\tau}$. This prevents shared normalization from distorting representations when the input statistics vary across lengths. The fused feature is computed as
\begin{equation}
\mathbf{F}_{\rm e}^{\tau}=\Phi_{\rm E}^{\rm OAF}\!\left(\mathbf{X}^{\tau},\mathbf{M};\,\Theta_{\rm E\text{-}shared}^{\rm OAF},\Theta_{\rm E\text{-}LN}^{\rm OAF,\tau}\right),
\end{equation}
where $\mathbf{F}_{\rm e}^{\tau}$ aggregates motion cues from $\mathbf{X}^{\tau}$ and map context from $\mathbf{M}$. We then extract AOI-specific features
\begin{equation}
\mathbf{F}_{\rm aoi}^{\tau}=\text{Extract}\!\left(\mathbf{F}_{\rm e}^{\tau};\,\text{AOI}\right).
\end{equation}
The decoder shares parameters $\Theta_{\rm D}^{\rm OAF}$ across all lengths to enforce consistent prediction logic, yielding the future trajectory prediction
\begin{equation}
\hat{\mathbf{Y}}^{\tau}=\Phi_{\rm D}^{\rm OAF}\!\left(\mathbf{F}_{\rm aoi}^{\tau};\,\Theta_{\rm D}^{\rm OAF}\right).
\end{equation}

\paragraph{Progressive knowledge distillation (PKD).}
While parameter sharing provides a unified predictor, extremely short histories still suffer from an intrinsic information deficit. PKD addresses this issue by progressively transferring motion-pattern knowledge from longer histories to shorter ones. Importantly, we distill knowledge using \emph{full-agent} features rather than AOI-only features, since inter-agent context is often crucial under partial observability. Specifically, from the fused encoder feature $\mathbf{F}_{\rm e}^{\tau}$, we extract features for all $N_{\rm a}$ agents:
\begin{equation}
\mathbf{F}_{\rm ag}^{\tau}=\text{Extract}\!\left(\mathbf{F}_{\rm e}^{\tau};\,N_{\rm a}\right).
\end{equation}
This full-context design improves robustness when the AOI history is truncated, as neighboring agents and map-conditioned interactions provide complementary cues for the missing states.

To enforce cross-length consistency without inducing severe feature mismatch, PKD performs \emph{adjacent-length} alignment along the hierarchy $\tau=1\rightarrow2\rightarrow\cdots\rightarrow H$. Unlike directly aligning very short inputs with the full-length history (which can be overly difficult and unstable), we align each student length $\tau$ with its adjacent teacher length $\tau+1$, which is naturally matched in spatiotemporal scope. Teacher features are detached to ensure one-way knowledge flow and to prevent short-history noise from corrupting reliable long-history representations. The alignment loss is defined as
\begin{equation}
\mathcal{L}_{\rm f}=\frac{1}{H-1}\sum_{\tau=1}^{H-1}\left\lVert \mathbf{F}_{\rm ag}^{\tau}-\text{detach}\!\left(\mathbf{F}_{\rm ag}^{\tau+1}\right)\right\rVert_{1}.
\end{equation}
This hierarchical, spatiotemporally matched alignment yields progressive distillation from long to short histories, reduces optimization difficulty compared with direct cross-level alignment, and preserves the benefits of full-agent context for handling practical missing-history scenarios.

\begin{figure*}[t!]
    \centering
    \includegraphics[width=0.9\textwidth]{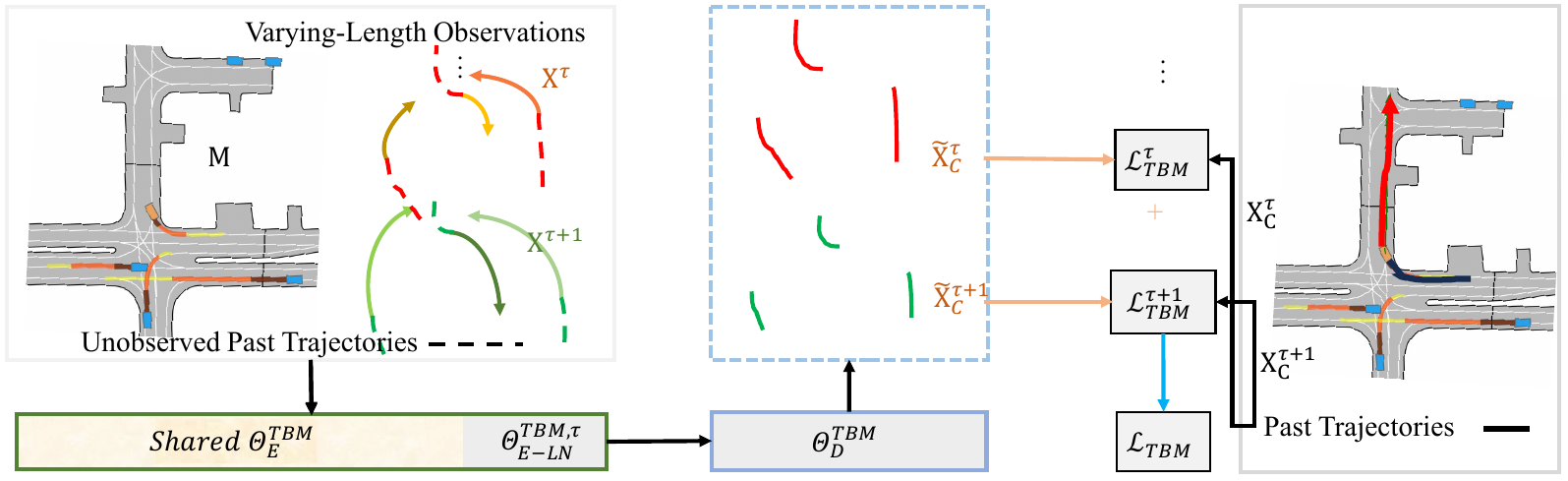}
    \caption{
    The TBM temporal backfilling module for unobserved past trajectories backfilling. 
    }
    \label{fig:main}
\end{figure*}

\subsection{Temporal Backfilling Module (TBM)}

TBM is an independent component that explicitly reconstructs the unobserved past for short truncated histories $\mathbf{X}^{\tau}$ ($\tau<H$). It targets a key limitation of purely feature-level adaptation in OAF (e.g., PKD): although distillation can align representations across observation lengths, it cannot recover \emph{trajectory-specific prior motion states} (e.g., pre-occlusion velocity trend or heading evolution) that are never observed in extremely short inputs. TBM addresses this information deficit by converting a short history into a completed full-length history, enabling OAF to operate on context-rich inputs. To maintain efficient adaptation across multiple short lengths, TBM follows the same intra-module parameter-sharing principle as OAF and adopts an encoder--decoder architecture specialized for history reconstruction.

\paragraph{Encoder with cross-length parameter sharing.}
Given a short history $\mathbf{X}^{\tau}$ and the HD map $\mathbf{M}$, the TBM encoder $\Phi_{\rm E}^{\rm TBM}$ extracts reconstruction-oriented features. Core layers (temporal encoding, inter-agent interaction, and map--trajectory fusion) share parameters $\Theta_{\rm E\text{-}shared}^{\rm TBM}$ across all $\tau$, while LayerNorm parameters are length-specific, denoted by $\Theta_{\rm E\text{-}LN}^{\rm TBM,\tau}$. This mirrors OAF’s normalization strategy and mitigates representation distortion caused by shared normalization under length-dependent input statistics. The encoded feature is
\begin{equation}
\mathbf{F}_{\rm rec}^{\tau}
= \Phi_{\rm E}^{\rm TBM}\!\left(\mathbf{X}^{\tau}, \mathbf{M};\, \Theta_{\rm E\text{-}shared}^{\rm TBM}, \Theta_{\rm E\text{-}LN}^{\rm TBM,\tau}\right),
\end{equation}
where $\mathbf{F}_{\rm rec}^{\tau}$ summarizes the scene context and the observed motion cues relevant for backfilling.

\paragraph{Decoder and temporal backfilling.}
The decoder $\Phi_{\rm D}^{\rm TBM}$ shares parameters $\Theta_{\rm D}^{\rm TBM}$ across all short lengths and predicts the missing prefix that precedes the observed segment. Concretely, TBM generates the missing historical segment
\begin{equation}
\tilde{\mathbf{X}}_{\rm C}^{\tau}
= \Phi_{\rm D}^{\rm TBM}\!\left(\mathbf{F}_{\rm rec}^{\tau};\, \Theta_{\rm D}^{\rm TBM}\right),
\end{equation}
where $\tilde{\mathbf{X}}_{\rm C}^{\tau}\in\mathbb{R}^{N\times (H-\tau)\Delta T \times C_a}$ corresponds to the $(H-\tau)\Delta T$ unobserved time steps in the standard window. The completed full-length history is then formed by concatenation along the temporal dimension:
\begin{equation}
\hat{\mathbf{X}}^{H} = \text{Concat}\!\left(\tilde{\mathbf{X}}_{\rm C}^{\tau}, \mathbf{X}^{\tau}\right).
\end{equation}
This operation preserves the original observed states in $\mathbf{X}^{\tau}$ and augments them with reconstructed prior context, yielding a temporally ordered and standardized history of length $T_{obs}$.

\paragraph{Role of TBM in the overall framework.}
By explicitly reconstructing missing historical segments, TBM provides trajectory-specific priors that implicit feature alignment alone cannot reliably infer under severe truncation. The resulting context-rich $\hat{\mathbf{X}}^{H}$ serves as an enriched input to OAF, allowing it to fully exploit PKD-trained motion priors for accurate forecasting even when the originally observed history is extremely short.

\subsection{Training Strategy}

\begin{figure*}[t!]
    \centering
    \includegraphics[width=0.9\textwidth]{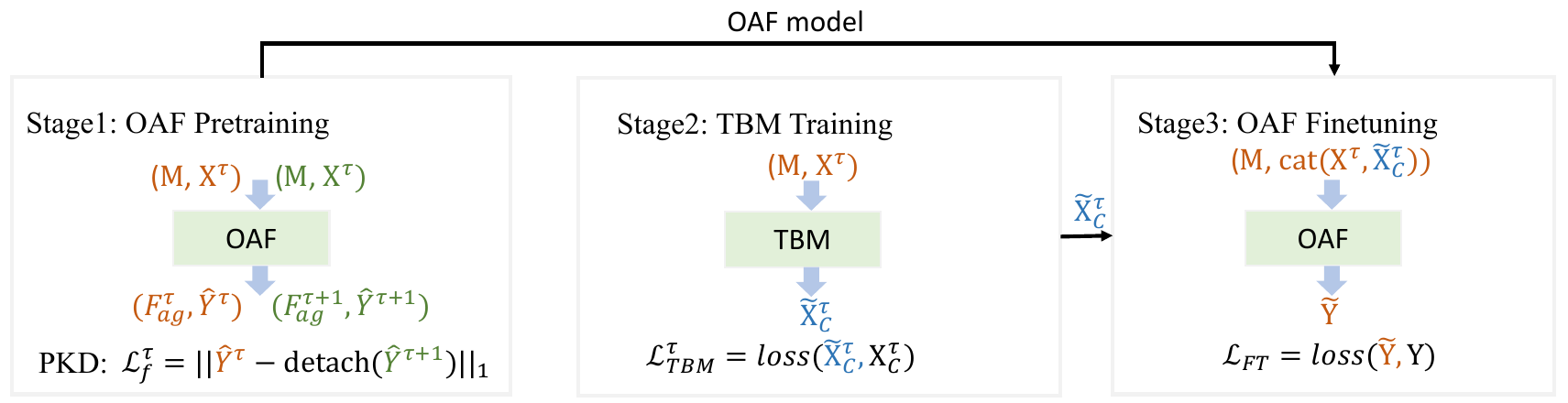}
    \caption{
    The overview of TaPD training strategy. 
    }
    \label{fig:main}
\end{figure*}

To fully exploit the complementarity between TBM and OAF while preserving reliable motion-prior learning and preventing adverse co-adaptation, we adopt a sequential three-stage training protocol. The protocol leverages the architectural decoupling of the two modules: each component is first optimized for its primary objective in isolation, and OAF is subsequently adapted to TBM-generated histories under controlled fine-tuning. This design yields robust performance across variables and extremely short observation lengths.

\subsubsection{Stage 1: Pre-train OAF}
In the first stage, we pre-train OAF solely on real driving data to establish strong forecasting priors and cross-length generalization. The inputs are native variable-length histories $\mathbf{X}^{\tau}$ (constructed by truncation or sliding windows) paired with the HD map $\mathbf{M}$; no reconstructed observations are used to avoid injecting reconstruction noise into motion-pattern learning. The objective jointly optimizes (i) accurate future trajectory prediction from variable-length inputs and (ii) progressive cross-length feature alignment via PKD. The overall loss is
\begin{equation}
\mathcal{L}_{\rm OAF} = \alpha \, \mathcal{L}_{\rm f} + \mathcal{L}_{\rm reg},
\end{equation}
where $\mathcal{L}_{\rm f}$ is the PKD feature-alignment loss and $\mathcal{L}_{\rm reg}$ aggregates prediction supervision across lengths:
\begin{equation}
\mathcal{L}_{\rm reg} = \sum_{\tau=1}^{H}\left(\mathcal{L}_{\rm reg}^{\tau} + \mathcal{L}_{\rm cls}^{\tau}\right).
\end{equation}
Here, $\mathcal{L}_{\rm reg}^{\tau}$ is a Smooth-$L_1$ regression loss between the predicted and ground-truth future trajectories, and $\mathcal{L}_{\rm cls}^{\tau}$ is a cross-entropy loss for motion-mode classification, which encourages consistent multimodal motion reasoning. The PKD weight $\alpha$ is scheduled by cosine annealing:
\begin{equation}
\alpha = \frac{1}{2}\left(1-\cos\left(\frac{e \cdot \pi}{E}\right)\right),
\end{equation}
\zh{where $e$ is the current epoch and $E$ is the total number of pre-training epochs.} This schedule mitigates early-stage instability by prioritizing trajectory supervision when representations are immature, and gradually increases the emphasis on feature alignment as OAF converges.

\subsubsection{Stage 2: Train TBM Independently}
In the second stage, we train TBM as a standalone reconstruction model to maximize the fidelity of backfilled histories, while keeping OAF completely isolated. TBM takes \zh{short histories $\mathbf{X}^{\tau}$ ($\tau<H$)} and the HD map $\mathbf{M}$ as inputs, and predicts the missing historical segment $\tilde{\mathbf{X}}^{\tau}_{C}$ such that the \zh{concatenated trajectory $\hat{\mathbf{X}}^{H}$} is temporally coherent and consistent with real motion dynamics. To ensure compatibility with OAF’s learning objectives, TBM uses a reconstruction loss aligned with OAF’s supervision:
\begin{equation}
\mathcal{L}_{\rm TBM} = \sum_{\tau=1}^{H-1}\left(\mathcal{L}_{\rm reg}^{\tau} + \mathcal{L}_{\rm cls}^{\tau}\right),
\end{equation}
where $\mathcal{L}_{\rm reg}^{\tau}$ is a Smooth-$L_1$ loss between the predicted and ground-truth missing segments, and $\mathcal{L}_{\rm cls}^{\tau}$ is a cross-entropy loss that enforces motion-mode consistency for the reconstructed segment. Training TBM independently prevents erroneous reconstruction artifacts from biasing OAF during representation learning.

\subsubsection{Stage 3: Freeze TBM and Fine-tune OAF}
In the final stage, we freeze TBM to preserve its reconstruction capability and fine-tune OAF to accommodate TBM-completed histories while retaining the motion priors learned from real data. For each training instance, if $\tau<H$, the truncated history $\mathbf{X}^{\tau}$ is first processed by the frozen TBM to produce a completed history $\hat{\mathbf{X}}^{H}$; if $\tau=H$, the native full-length history $\mathbf{X}^{H}$ is used directly. OAF is then trained on these (completed or native) full-length inputs.

During fine-tuning, we optimize only the trajectory prediction objective and omit the PKD alignment term to avoid overwriting the motion representations acquired in Stage~1. The fine-tuning loss is therefore
\begin{equation}
\mathcal{L}_{\rm FT}=\tilde{\mathcal{L}}_{\rm reg}^{H},
\end{equation}
where $\tilde{\mathcal{L}}_{\rm reg}^{H}$ denotes the standard prediction loss computed from OAF outputs given the (TBM-completed or native) full-length history. This sequential protocol yields a final model that combines reliable forecasting priors from real trajectories with explicit history completion, leading to robust performance under variable-length and extremely short observations.

\section{Experiments}

\subsection{Experimental Settings}
\paragraph{Datasets.}
We conduct experiments on two established motion forecasting benchmarks, Argoverse~1~\cite{argoverse} and Argoverse~2~\cite{argoverse2}. 
Argoverse~1 comprises 323{,}557 real-world driving scenarios collected in the Miami and Pittsburgh metropolitan areas. Each scenario is sampled at 10\,Hz and spans 5\,s, where the first 2\,s (20 frames) are provided as observed history and the remaining 3\,s (30 frames) are used as the prediction horizon. 
Argoverse~2 further increases both scale and diversity, containing 250{,}000 scenarios from six cities. Each sequence is recorded at 10\,Hz over 11\,s; following the official protocol, we use the first 5\,s (50 frames) as input and predict the subsequent 6\,s (60 frames).

\paragraph{Evaluation Metrics.}
We report three standard metrics for multimodal trajectory forecasting: minimum Average Displacement Error (minADE$_K$), minimum Final Displacement Error (minFDE$_K$), and Miss Rate (MR$_K$). 
Given a set of $K$ predicted trajectories $\{\hat{\mathbf{Y}}^{(k)}\}_{k=1}^{K}$ and the ground-truth future trajectory $\mathbf{Y}$, we define
\begin{equation}
\mathrm{ADE}\big(\hat{\mathbf{Y}}^{(k)},\mathbf{Y}\big)=\frac{1}{T}\sum_{t=1}^{T}\left\lVert \hat{\mathbf{y}}^{(k)}_{t}-\mathbf{y}_{t}\right\rVert_{2},
\end{equation}
\begin{equation}
\mathrm{FDE}\big(\hat{\mathbf{Y}}^{(k)},\mathbf{Y}\big)=\left\lVert \hat{\mathbf{y}}^{(k)}_{T}-\mathbf{y}_{T}\right\rVert_{2},
\end{equation}
where $T$ denotes the number of predicted time steps and the $\ell_2$ distance is measured in meters. 
The multimodal errors are then computed in a best-of-$K$ manner:
\begin{equation}
\mathrm{minADE}_{K}=\min_{k\in\{1,\dots,K\}}  
\mathrm{ADE}\big(\hat{\mathbf{Y}}^{(k)},\mathbf{Y}\big), 
\end{equation}
\begin{equation}
\mathrm{minFDE}_{K}=\min_{k\in\{1,\dots,K\}}  \mathrm{FDE}\big(\hat{\mathbf{Y}}^{(k)},\mathbf{Y}\big).
\end{equation}
Following prior work, the miss rate is defined as
\begin{equation}
\mathrm{MR}_{K}=\mathbb{P}\big(\mathrm{minFDE}_{K} > 2\,\mathrm{m}\big),
\end{equation}
i.e., the percentage of samples whose best final displacement exceeds 2\,m. 
Unless otherwise specified, we report results for $K\in\{1,6\}$.

\paragraph{Backbone and Baselines.}
We instantiate our approach on top of DeMo~\cite{demo} and benchmark against representative baselines designed for variable observation lengths, including DTO~\cite{howmany}, FLN~\cite{FLN}, LaKD~\cite{lakd}, and CLLS~\cite{adapting}. 
We additionally report two reference settings for clarity. 
\textbf{Ori} denotes a model trained exclusively with the standard observation length and then directly evaluated under observation-length shift. 
\textbf{IT} (Isolated Training) trains a separate model for each observation length and evaluates each model on its matched length.

\paragraph{Implementation Details.}
We construct variable-length histories by truncating the standard full history of length $T_{obs}$ \zh{by removing several intervals, each of duration $\Delta T$.}
Concretely, we set $\Delta T{=}10$ and $T_{obs}{=}50$ time steps on Argoverse~2, and $\Delta T{=}5$ and $T_{obs}{=}20$ on Argoverse~1. 
DeMo~\cite{demo} is used as the backbone. While the original DeMo includes the RealMotion module~\cite{relmotion}, we remove RealMotion and retain only the core forecasting architecture to obtain a clean and reproducible baseline for our extensions.
All models are trained for 60 epochs using AdamW~\cite{adamw} with an initial learning rate of $6{\times}10^{-4}$, weight decay of 0.01, and a batch size of 16. 
We adopt an agent-centric coordinate system and include map/scene elements within a 150\,m radius centered at each target agent. 
All experiments are conducted on two NVIDIA A40 GPUs.

\subsection{Performance on Variable-Length Histories}
\input{table/table_compare}

Table~\ref{tab:av2_av1_compare} summarizes the variable-length forecasting results on the Argoverse~2 and Argoverse~1 validation sets. 
As expected, the standard-length model (\textbf{Ori}) exhibits pronounced performance degradation when the observation history shortens, indicating that naive fixed-window training is not robust to observation-length shift. For instance, on Argoverse~2, \textbf{Ori} deteriorates from 0.658/1.278 at 50Ts to 0.861/1.533 at 10Ts (minADE$_6$/minFDE$_6$), and on Argoverse~1 it drops from 0.606/1.003 at 20Ts to 0.781/1.267 at 5Ts. 
While \textbf{IT} (Isolated Training) improves short-history performance by training length-specific models (e.g., 0.675/1.318 on Argoverse~2 at 10Ts and 0.669/1.078 on Argoverse~1 at 5Ts), its gains quickly diminish as the history approaches the standard length (nearly identical to \textbf{Ori} at 50Ts/20Ts). 
\zh{Moreover, IT is impractical in real-world deployments because it requires maintaining multiple models.}

Among dedicated length-adaptive baselines (\textbf{DTO}, \textbf{FLN}, \textbf{LaKD}, \textbf{CLLS}), our \netName{} consistently achieves the best performance across \emph{all} observation lengths on both datasets. 
The improvements are most evident under extremely short histories: on Argoverse~2 at 10Ts, \netName{} reduces the error from 0.861/1.533 (\textbf{Ori}) to 0.617/1.203, and also surpasses the strongest adaptive baseline (0.641/1.258 by \textbf{CLLS}); on Argoverse~1 at 5Ts, \netName{} achieves 0.608/0.974 compared to 0.781/1.267 (\textbf{Ori}) and 0.634/0.998 (\textbf{CLLS}). 
More importantly, \netName{} substantially narrows the short-to-full performance gap: the minFDE$_6$ gap between 10Ts and 50Ts on Argoverse~2 is reduced from 0.255 (\textbf{Ori}) to 0.050, and the gap between 5Ts and 20Ts on Argoverse~1 decreases from 0.264 to 0.060. 
This trend supports the proposed dual-module design: TBM explicitly compensates for missing historical context, while PKD in OAF encourages representation consistency across lengths. 
The consistent gains on both Argoverse~1 and Argoverse~2 further demonstrate the robustness and generality of \netName{} under variable-length observations.

\subsection{Comparison with State of the Art}
\input{table/table_av2_single_val}

\input{table/table_av1}

Although \netName{} is tailored for observation-adaptive forecasting under variable-length histories, it also yields competitive---and in several cases state-of-the-art---performance under the conventional fixed-length protocol. Tables~\ref{tab:av2_single} and \ref{tab:table_av1} report single-agent results on the Argoverse~2~\cite{argoverse2} and Argoverse~1~\cite{argoverse} validation sets, respectively.

\textbf{Argoverse~2.} As shown in Table~\ref{tab:av2_single}, \netName{} achieves the best overall accuracy. It attains the lowest $\textit{minADE}_{6}$ of 0.59 and the lowest $\textit{MR}_{6}$ of 0.13, while matching the best $\textit{minFDE}_{6}$ of 1.15 (tied with LITNT~\cite{litnt}). Compared with the DeMo backbone~\cite{demo}, \netName{} reduces $\textit{minFDE}_{6}$ from 1.27 to 1.15 (\textbf{9.4\%} relative improvement) and $\textit{minADE}_{6}$ from 0.67 to 0.59 (\textbf{11.9\%}), and further lowers $\textit{MR}_{6}$ from 0.15 to 0.13. These gains suggest that the proposed training scheme not only improves robustness to observation-length shift, but also enhances the quality of multimodal trajectory hypotheses even when evaluated with the standard full-history input.

\textbf{Argoverse~1.} Table~\ref{tab:table_av1} shows that \netName{} achieves a new best $\textit{minADE}_{6}$ of 0.57 and the best $\textit{MR}_{6}$ of 0.07 (tied with HPNet~\cite{hpnet}). For $\textit{minFDE}_{6}$, \netName{} remains highly competitive at 0.91, ranking second behind HPNet (0.87) while improving over DeMo (1.00$\rightarrow$0.91). Overall, \netName{} consistently strengthens the DeMo backbone on both datasets and maintains state-of-the-art fixed-length performance, despite being explicitly designed to handle variable-length observations.

\subsection{Ablation Study}
\input{table/table_ab_main}

\input{table/table_ab_combination}

\paragraph{Effects of components.}
Table~\ref{tab:ab_main} summarizes an ablation study on the Argoverse~2 validation set to quantify the contribution of each core component. The first row is the baseline without any of our proposed mechanisms (i.e., no parameter sharing, no PKD, and no TBM), which yields $\textit{minADE}_6/\textit{minFDE}_6=0.675/1.318$ at 10Ts and $0.658/1.278$ at 50Ts.

Adding only the PS mechanism consistently improves performance across all observation lengths by unifying the feature extraction/prediction logic and avoiding redundant length-specific optimization. For instance, PS improves 10Ts from $0.675/1.318$ to $0.650/1.289$ and 50Ts from $0.658/1.278$ to $0.616/1.211$. 

Introducing PKD on top of PS yields further gains, with a larger effect under shorter histories. This supports our motivation that progressive distillation can effectively transfer spatiotemporal motion priors from longer observations to shorter ones and partially compensate for missing temporal context. Concretely, PS+PKD improves 10Ts from $0.650/1.289$ to $0.621/1.216$ and 50Ts from $0.616/1.211$ to $0.601/1.172$.

Finally, the full model (PS+PKD+TBM) achieves the best results across all lengths. The additional gains are most evident for short inputs, where the information deficit is most severe: compared with PS+PKD, TBM further reduces 10Ts from $0.621/1.216$ to $0.617/1.203$ and 20Ts from $0.610/1.200$ to $0.603/1.167$. Overall, the ablation confirms that PS, PKD, and TBM play complementary roles: PS establishes an efficient unified predictor across lengths, PKD strengthens cross-length knowledge transfer, and TBM explicitly restores trajectory-specific historical context that feature-level alignment alone cannot reliably recover.

\paragraph{Effects of length combination.}

Table~\ref{tab:combination} investigates how different training-length combinations affect variable-length generalization on Argoverse~2. Training on longer histories only (DeMo\_TaPD-3 with 30/40/50) performs competitively on the included lengths and improves over isolated training at 30--50Ts (e.g., 50Ts: $0.605/1.174$ vs.\ $0.658/1.278$ for DeMo\_IT). However, its performance degrades substantially on unseen shorter lengths, most notably at 20Ts ($0.627/1.224$) and 10Ts ($0.671/1.313$), indicating limited extrapolation toward short-history regimes when such lengths are absent during training.

Expanding the training set to include 20Ts (DeMo\_TaPD-4 with 20/30/40/50) further improves performance on 20--50Ts (e.g., 20Ts: $0.606/1.188$; 50Ts: $0.603/1.162$), but a clear gap remains at the unseen 10Ts case ($0.651/1.274$), again highlighting the difficulty of generalizing to extremely short histories without direct exposure.

In contrast, DeMo\_TaPD-5, trained with the full length spectrum (10/20/30/40/50), achieves the strongest and most consistent results across all observation lengths, improving over DeMo\_IT at every time step (e.g., 10Ts: $0.617/1.203$ vs.\ $0.675/1.318$; 50Ts: $0.599/1.153$ vs.\ $0.658/1.278$). While incorporating more lengths increases training cost, the results demonstrate that broad length coverage is crucial for robust observation-adaptive forecasting, particularly for extreme short-history inputs.

\subsection{Plug-and-Play Evaluation}
\input{table/table_av2_moremodel}

\input{table/table_HRM}

To verify the plug-and-play property of \netName{}, we integrate our framework into a representative backbone, HiVT~\cite{hivt}, and evaluate it under variable-length observations on the Argoverse~1 single-agent validation set (Table~\ref{tab:moremodel}). The original HiVT model (\textbf{HiVT\_Ori}) suffers a substantial performance drop under short histories, e.g., $0.943/1.540$ at 5Ts. Isolated Training (\textbf{HiVT\_IT}) mitigates this issue but provides limited improvement at longer lengths and requires training separate models. In contrast, \textbf{HiVT\_TaPD} consistently achieves the best results at all observation lengths, reducing errors from $0.781/1.224$ to $0.704/1.083$ at 5Ts and from $0.702/1.070$ to $0.660/0.964$ at 20Ts. Moreover, our method outperforms the length-adaptive baseline FLN across the board (e.g., 5Ts: $0.754/1.162 \rightarrow 0.704/1.083$; 20Ts: $0.681/1.028 \rightarrow 0.660/0.964$). These results demonstrate that \netName{} can be seamlessly incorporated into existing forecasting pipelines and reliably improves variable-length performance without architectural overhauls.

\subsection{Performance of TBM}

Table~\ref{tab:table_HRM} reports the backfilling accuracy of TBM on Argoverse~2 and Argoverse~1 validation sets. A clear monotonic trend is observed: as the available observation length increases, both $\textit{minADE}_6$ and $\textit{minFDE}_6$ decrease consistently across datasets. On Argoverse~2, the reconstruction error drops from $0.382/0.723$ at 10Ts to $0.097/0.160$ at 40Ts, while on Argoverse~1 it decreases from $0.316/0.307$ at 5Ts to $0.195/0.111$ at 15Ts. This behavior indicates that TBM effectively exploits additional temporal evidence to recover missing history more accurately, and remains robust under varying degrees of truncation. In turn, the improved backfilled histories provide richer context for OAF, which is particularly beneficial in the extremely short-history regime.

\zh{\subsection{Model Efficiency Analysis}}
\input{table/table_efficiency}

\zh{Table~\ref{tab:model_efficiency} presents the model efficiency of the DeMo backbone and the proposed TaPD on the Argoverse 2 dataset. Since TaPD's inference cost on full-length input is equal to that of DeMo\_IT, we report results only for incomplete input lengths. The results show that DeMo\_IT is insensitive to input length across all three metrics, whereas TaPD is insensitive to input length with respect to parameters and latency but incurs increased FLOPs as input length decreases. The largest inference overhead of TaPD occurs at an input length of 10Ts, where it increases over DeMo\_IT by 3.400 M parameters, 0.535 G FLOPs, and 65.198 ms latency. Despite a modest increase in inference overhead, our method achieves real-time inference even under the heaviest workload, with a latency of 88.568 ms, thereby demonstrating its lightweight design.}

\subsection{Qualitative Comparisons}

Fig.~\ref{fig:visual} provides qualitative comparisons under extremely short observations (10 time steps). 
In the first two rows, DeMo\_IT tends to produce less reliable multimodal hypotheses and may overshoot the ground-truth future trajectory, suggesting that training isolated models per length does not sufficiently capture informative motion cues from a very limited history. 
In contrast, our DeMo\_OAF and DeMo\_TaPD generate trajectories with noticeably improved coverage of the ground truth, indicating stronger motion representation under short inputs. 
Moreover, DeMo\_TaPD exhibits \zh{accurate} backfilling of the unobserved history (blue vs.\ orange), which provides a better-conditioned context for forecasting.

In the last two rows, DeMo\_IT shows a substantial deviation from the ground-truth future (red), while DeMo\_OAF partially corrects the trend but still exhibits residual bias. 
Our DeMo\_TaPD achieves the closest alignment: it reconstructs the missing historical segment \zh{faithfully} and yields multimodal predictions that better match both the trajectory trend and the distribution of plausible futures.
These cases qualitatively support the roles of TBM in compensating for missing context and of PKD in promoting cross-length consistency.

\input{figure/fig_visual}

\section{Conclusion}
This paper addresses variable-length motion forecasting in autonomous driving, where the available observation history can be arbitrarily short due to occlusions, limited sensing range, or late agent entry. We propose \netName{}, a unified framework that combines two complementary modules: an observation-adaptive forecaster and a temporal backfilling module. The observation-adaptive forecaster provides a single predictor for all observation lengths via a parameter-sharing design, and further improves short-history robustness through progressive knowledge distillation, which transfers spatiotemporal motion priors from longer trajectories to shorter ones. The temporal backfilling module complements the observation-adaptive forecaster by explicitly reconstructing missing historical segments for truncated inputs, supplying trajectory-specific prior states that cannot be reliably recovered by feature-level alignment alone. Extensive experiments on Argoverse~1 and Argoverse~2 demonstrate that \netName{} consistently improves $\textit{minADE}_6$ and $\textit{minFDE}_6$ across all observation lengths, markedly narrowing the performance gap between short and full histories while also achieving competitive performance on standard fixed-length benchmarks.

In future work, we plan to extend \netName{} by generalizing the current alignment strategy from feature-level matching to a multi-granularity formulation that jointly enforces representation-, intent-, and motion-pattern consistency, enabling more structured adaptation across observation lengths. We expect these extensions to further improve robustness in real-world driving environments with dense interactions and severe partial observability.


{
\bibliographystyle{IEEEtran} 
\bibliography{cite}
}

\vfill

\end{document}

%% file: figure/fig_main.tex
\begin{figure*}[t!]
    \centering
    \includegraphics[width=0.9\textwidth]{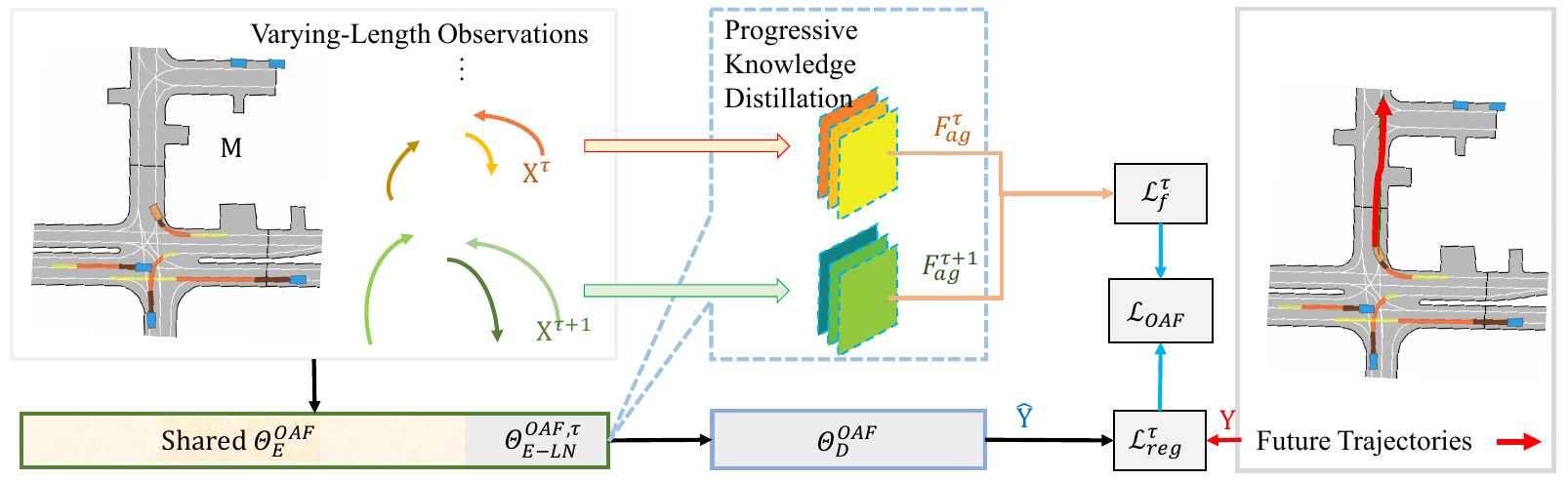}
    \caption{
    The OAF pipeline with parameter sharing and progressive knowledge distillation.}
    \label{fig:main}
\end{figure*}

%% file: table/table_compare.tex
\begin{table*}[t!]
\caption{Variable-length trajectory prediction comparison on Argoverse 2 (left) and Argoverse 1 (right) validation sets. Best results are in \textbf{bold} and second best are \underline{underlined}.  Ts denotes the time steps of $T_{\tau }, a.k.a., \tau \Delta T$.}
\label{tab:av2_av1_compare}
\centering
\resizebox{0.95\textwidth}{!}{
\begin{tabular}{l ccccc | cccc}
\toprule[1.5pt]
\multirow{2}{*}{Method} & 
\multicolumn{5}{c}{Argoverse 2 ($\textit{minADE}_{6}$/$\textit{minFDE}_{6}$)} & 
\multicolumn{4}{c}{Argoverse 1 ($\textit{minADE}_{6}$/$\textit{minFDE}_{6}$)} \\
\cmidrule(lr){2-6} \cmidrule(lr){7-10} 
& 10Ts & 20Ts & 30Ts & 40Ts & 50Ts & 5Ts & 10Ts & 15Ts & 20Ts \\
\midrule[1pt] 
DeMo\_Ori
& 0.861/1.533 & 0.700/1.358 & 0.671/1.306 & 0.662/1.288 & 0.658/1.278
& 0.781/1.267 & 0.662/1.087 & 0.624/1.011 & 0.606/1.003 \\
DeMo\_IT 
& 0.675/1.318 & 0.661/1.296 & 0.660/1.293 & 0.659/1.287 & 0.658/1.278
& 0.669/1.078 & 0.634/1.031 & 0.612/0.988 & 0.606/1.003 \\
DeMo\_DTO
& 0.672/1.307 & 0.658/1.291 & 0.650/1.279 & 0.647/1.268 & 0.645/1.265
& 0.662/1.064 & 0.628/1.025 & 0.605/0.991 & 0.599/1.010 \\
DeMo\_FLN 
& 0.651/1.262 & 0.644/1.258 & 0.637/1.254 & 0.628/1.238 & 0.621/1.231
& 0.646/1.043 & 0.607/0.994 & 0.599/0.974 & 0.592/0.957 \\
DeMo\_LaKD
& \underline{0.639}/1.262 & \underline{0.627}/1.251 & \underline{0.620}/1.243 & 0.617/1.236 & 0.617/1.232
& \underline{0.631}/1.008 & 0.593/0.976 & 0.584/0.933 & 0.581/0.929 \\
DeMo\_CLLS
& 0.641/\underline{1.258} & 0.630/\underline{1.249} & 0.623/\underline{1.234} & \underline{0.614}/\underline{1.225} & \underline{0.615}/\underline{1.223}
& 0.634/\underline{0.998} & \underline{0.587}/\underline{0.959} & \underline{0.580}/\underline{0.919} & \underline{0.579}/\underline{0.922} \\
DeMo\_TaPD(Ours)
& \bf0.617/\bf1.203 & \bf0.603/\bf1.167 & \bf0.599/\bf1.157 & \bf0.599/\bf1.155 & \bf0.599/\bf1.153
& \bf0.608/\bf0.974 & \bf0.574/\bf0.915 & \bf0.573/\bf0.914 & \bf0.573/\bf0.914 \\
\bottomrule[1.5pt]
\end{tabular}
}
\end{table*}

%% file: table/table_av2_single_val.tex
\begin{table} [ht!]
    \caption{Performance comparison on {\it Argoverse 2 validation set}. "-" denotes that this result was not reported in their paper.}
    \label{tab:av2_single}
    \centering
    \begin{tabular}{l|ccc}
       \toprule[1.5pt]
        $\textbf{Method}$ & $\textit{minFDE}_{6}$ & $\textit{minADE}_{6}$ & $\textit{MR}_{6}$ \\
        [1.5pt]\hline\noalign{\vskip 2pt}
        LGT \cite{LGT}                 & 1.93 & 1.02 & 0.30 \\
        SIMPL~\cite{simpl}             & 1.45 & 0.78 & - \\
        forecast-mae~\cite{Forecast-mae} & 1.41 & 0.80 & 0.18 \\
        SeNeVA \cite{seNeVA}           & 1.32 & 0.71 & 0.17 \\
        Realmotion~\cite{relmotion}    & 1.31 & 0.66 & 0.15 \\
        QCNet~\cite{qcnet}             & 1.27 & 0.73 & 0.16 \\
        DeMo~\cite{demo}               & 1.27 & 0.67 & 0.15 \\
        LANet \cite{lanet}             & 1.26 & 0.72 & 0.16 \\
        Polaris \cite{Polaris}         & 1.21 & \underline{0.63} & \underline{0.14} \\
        FutureNet-LoF~\cite{futurenet} & 1.19 & 0.71 & \underline{0.14} \\
        CG-Net \cite{CG-Net}           & \underline{1.18} & 0.70 & \underline{0.14} \\
        LITNT \cite{litnt}             & \bf 1.15 & 0.70 & 0.15 \\
        \hline\noalign{\vskip 2pt}
        \bf\netName~(Ours)             & \bf 1.15 & \bf 0.59 & \bf 0.13 \\
        [1.5pt] \hline\noalign{\vskip 2pt}
    \end{tabular}
\end{table}

%% file: table/table_av1.tex
\begin{table} [ht!]
    \caption{Performance comparison on {\it Argoverse 1 validation set}. "-" denotes that this result was not reported in their paper.}
    \label{tab:table_av1}
    \centering
    {\begin{tabular}{l|ccc}
       \toprule[1.5pt]
        $\textbf{Method}$ & $\textit{minADE}_{6}$ & $\textit{minFDE}_{6}$ & $\textit{MR}_{6}$\\[1.5pt]\hline\noalign{\vskip 2pt}
        
        LTP~\cite{ltp}                   & 0.78 & 1.07 & -\\ 
        LaneRCNN~\cite{lanercnn}         & 0.77 & 1.19 & \underline{0.08}\\
        TPCN~\cite{tpcn}                   & 0.73 & 1.15 & 0.11\\
        DenseTNT~\cite{densetnt}               & 0.73 & 1.05 & 0.10\\
        TNT~\cite{tnt}               & 0.73 & 1.29 & 0.09\\
        mmTransformer~\cite{mmTransformer}               & 0.71 & 1.15 & 0.11\\
        LaneGCN~\cite{LaneGCN}               & 0.71 & 1.08 & -\\
        SSL-Lanes~\cite{ssl}          & 0.70 & 1.01 & 0.09 \\
        PAGA~\cite{paga}               & 0.69 & 1.02 & -\\
        DSP~\cite{DSP}               & 0.69 & 0.98 & 0.09\\
        FRM~\cite{frm}                & 0.68 & 0.99 & - \\
        ADAPT~\cite{adapt}               & 0.67 & 0.95 & \underline{0.08}\\
        SIMPL~\cite{simpl}            & 0.66 & 0.95 & \underline{0.08}\\
        HiVT~\cite{hivt}               & 0.66 & 0.96 & 0.09\\
        R-Pred~\cite{R-Pred}               & 0.66 & 0.95 & 0.09\\   
        HPNet~\cite{hpnet}      & 0.64 & \bf 0.87 & \bf 0.07\\
        DeMo~\cite{demo}       & \underline{0.61}   & 1.00  & 0.09 \\
        \hline\noalign{\vskip 2pt}
        \bf\netName~(Ours)       & \bf 0.57 & \underline{0.91} & \bf 0.07\\
        
        \bottomrule[1.5pt]
    \end{tabular}}
\end{table}

%% file: table/table_ab_main.tex
\begin{table*}[t!]
    \caption{Ablation study on the core components of~\netName~on the {\it Argoverse 2 validation set}. PS denotes parameter-sharing mechanism.}
    \label{tab:ab_main}
    \centering
    \begin{tabular}{c c c | c c c c c} 
       \toprule[1.5pt]
       \multirow{2}{*}{PS(OAF)} & \multirow{2}{*}{PKD(OAF)} & \multirow{2}{*}{TBM} & \multicolumn{5}{c}{$\textit{minADE}_6$ / $\textit{minFDE}_6$} \\
       \cmidrule(r){4-8} 
       & & & 10Ts & 20Ts & 30Ts & 40Ts & 50Ts \\
       \midrule 
       
       & & & 0.675/1.318 & 0.661/1.296 & 0.660/1.293 & 0.659/1.287 & 0.658/1.278 \\
       
       \checkmark & & & 0.650/1.289 & 0.637/1.256 & 0.629/1.238 & 0.624/1.226 & 0.616/1.211 \\
       
       \checkmark & \checkmark &  & 0.621/1.216 & 0.610/1.200 & 0.608/1.196 & 0.604/1.185 & 0.601/1.172 \\

       \checkmark & \checkmark & \checkmark & \bf0.617/\bf1.203 & \bf0.603/\bf1.167 & \bf0.599/\bf1.157 & \bf0.599/\bf1.155 & \bf0.599/\bf1.153 \\
       
       \bottomrule[1.5pt]
    \end{tabular}
\end{table*}

%% file: table/table_ab_combination.tex
\begin{table*}[t]
    \caption{Ablation study of different length combinations of observation lengths on the \textit{Argoverse 2 single-agent validation set}.}
    \label{tab:combination}
    \centering
    \begin{tabular}{c c ccccc}
       \toprule[1.5pt]   
        \multirow{2}{*}{Method} & \multirow{2}{*}{Combination} & 
         \multicolumn{5}{c}{$\textit{minADE}_{6}$ / $\textit{minFDE}_{6}$} \\
         \cmidrule(r){3-7}
         & &10Ts & 20Ts & 30Ts & 40Ts & 50Ts \\
         [1.5pt]\hline\noalign{\vskip 2pt}

        DeMo\_IT  & -                & 0.675/1.318 & 0.661/1.296
                                 & 0.660/1.293 & 0.659/1.287 & 0.658/1.278 \\ 
        DeMo\_TaPD-3  & 30/40/50      & 0.671/1.313 & 0.627/1.224 & 0.609/1.201 
                                 & 0.608/1.195 & 0.605/1.174 \\ 
        DeMo\_TaPD-4 & 20/30/40/50    & 0.651/1.274  & 0.606/1.188  & 0.605/1.182
                                 & 0.605/1.178 & 0.603/1.162  \\ 
        DeMo\_TaPD-5 & 10/20/30/40/50 & \bf0.617/\bf1.203 & \bf0.603/\bf1.167 &                                            \bf0.599/\bf1.157
                                         & \bf0.599/\bf1.155 & \bf0.599/\bf1.153 \\ 
        
        \bottomrule[1.5pt]
    \end{tabular}

\end{table*}

%% file: table/table_av2_moremodel.tex
\begin{table}[t!]
\caption{Performance Comparison of TaPD on HiVT on {\it Argoverse 1 single-agent validation set}.}
\label{tab:moremodel}
\centering
\setlength{\tabcolsep}{3pt} 
\footnotesize 
\begin{tabular}{l cccc}
\toprule[1.5pt]
\multirow{2}{*}{Method} & 
\multicolumn{4}{c}{$\textit{minADE}_{6}$ / $\textit{minFDE}_{6}$} \\
\cmidrule(lr){2-5} 
& 5Ts & 10Ts & 15Ts & 20Ts \\
[1.5pt]\hline\noalign{\vskip 2pt}
HiVT\_Ori & 0.943/1.540 & 0.768/1.201 & 0.732/1.124 & 0.702/1.070
\\ 
HiVT\_IT & 0.781/1.224 & 0.738/1.121 & 0.711/1.083 & 0.702/1.070
\\
HiVT\_FLN & 0.754/1.162 & 0.723/1.089 & 0.701/1.053 & 0.681/1.028
\\
HiVT\_TaPD(Ours) & \bf0.704/\bf1.083 & \bf0.662/\bf1.012 & \bf0.674/\bf1.003 & \bf0.660/\bf0.964
\\
[1.5pt]\hline\noalign{\vskip 2pt}
\end{tabular}
\end{table}

%% file: table/table_HRM.tex
\begin{table}[t!]
\caption{Variable-length trajectory prediction performance of TBM on Argoverse 2 (upper) and Argoverse 1 (lower) validation sets.}
\label{tab:table_HRM}
\centering
\setlength{\tabcolsep}{5pt}
\footnotesize

\begin{tabular*}{\linewidth}{@{\extracolsep{\fill}} lcccc}
\toprule[1.5pt]
\multicolumn{5}{c}{Argoverse 2 ($\textit{minADE}_{6}$/$\textit{minFDE}_{6}$)} \\
\midrule[0.8pt]
Module & 10Ts & 20Ts & 30Ts & 40Ts \\
\midrule[0.8pt]
TBM & 0.382/0.723 & 0.275/0.501 & 0.185/0.316 & 0.097/0.160 \\
\bottomrule[1pt]
\end{tabular*}

\vspace{0.4em}

\begin{tabular*}{\linewidth}{@{\extracolsep{\fill}} lccc}
\toprule[1pt]
\multicolumn{4}{c}{Argoverse 1 ($\textit{minADE}_{6}$/$\textit{minFDE}_{6}$)} \\
\midrule[0.8pt]
Module & 5Ts & 10Ts & 15Ts \\
\midrule[0.8pt]
TBM & 0.316/0.307 & 0.250/0.185 & 0.195/0.111 \\
\bottomrule[1.5pt]
\end{tabular*}

\end{table}

%% file: table/table_efficiency.tex

\begin{table}[t!]
\centering
\caption{Parameters (M), FLOPs (G), and latency (ms) under different input lengths.}
\label{tab:model_efficiency}
\footnotesize
\setlength{\tabcolsep}{5pt}

\begin{tabular}{c ccc ccc}
\toprule
\multirow{2}{*}{Len} &
\multicolumn{3}{c}{DeMo\_IT} &
\multicolumn{3}{c}{DeMo\_TaPD} \\
\cmidrule(lr){2-4}\cmidrule(lr){5-7}
& Params$\downarrow$ & FLOPs$\downarrow$ & Lat.$\downarrow$
& Params$\downarrow$ & FLOPs$\downarrow$ & Lat.$\downarrow$ \\
\midrule
10Ts & 3.431 & 0.678 & 23.370 & 6.831 & 1.213 & 88.568 \\
20Ts & 3.431 & 0.680 & 23.578 & 6.815 & 1.140 & 80.626 \\
30Ts & 3.431 & 0.683 & 23.878 & 6.800 & 1.066 & 84.238 \\
40Ts & 3.431 & 0.685 & 23.760 & 6.784 & 0.992 & 86.447 \\
\bottomrule
\end{tabular}
\end{table}

%% file: figure/fig_visual.tex
\begin{figure*}[htbp]
    \centering
    \setlength{\fboxsep}{0pt}
    \setlength{\fboxrule}{0pt}

    \captionsetup[subfigure]{position=bottom,justification=centering,skip=0.2em}

    \begin{subfigure}[b]{0.27\textwidth}
        \centering
        \includegraphics[width=\linewidth]{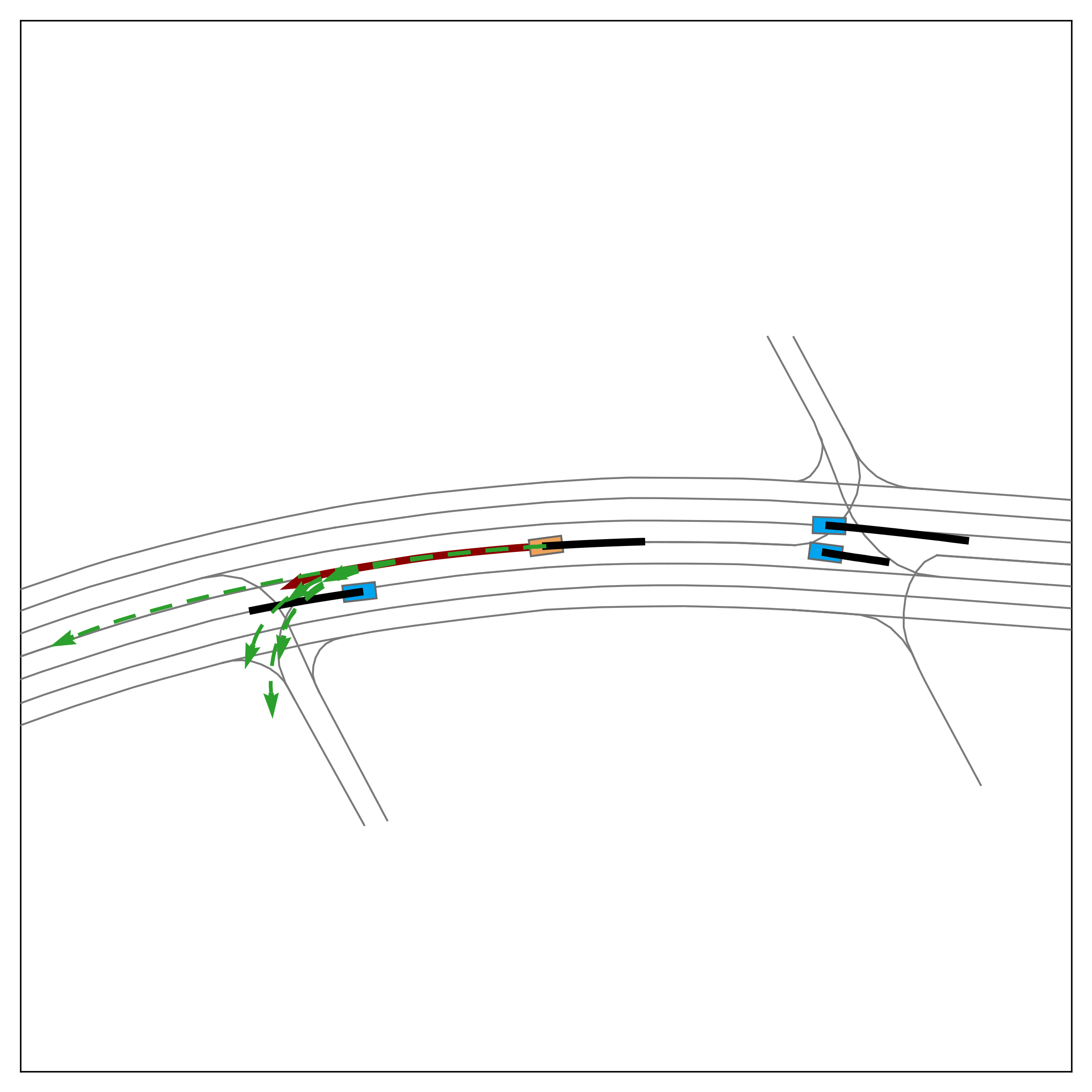}\par\vspace{0.5em}
        \includegraphics[width=\linewidth]{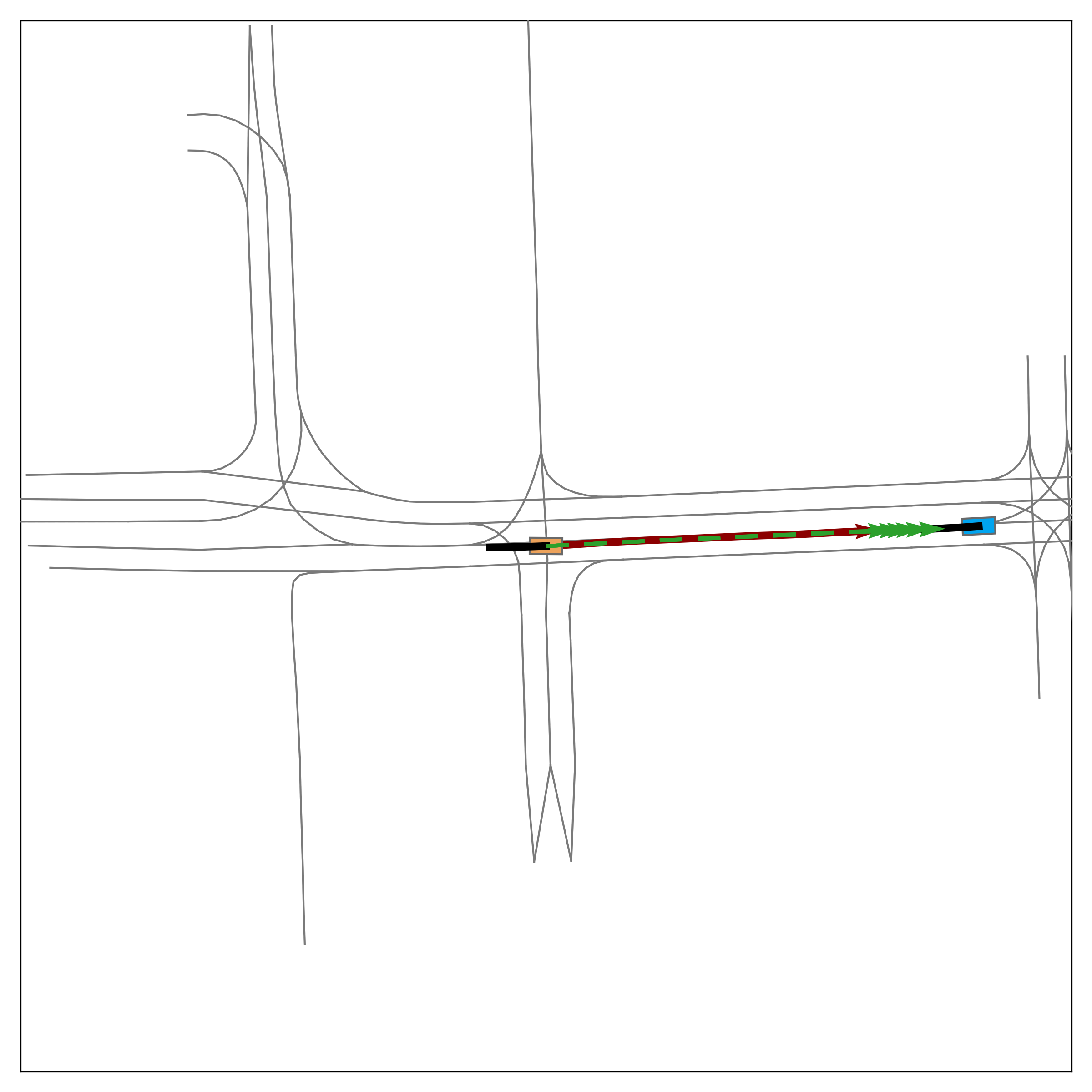}\par\vspace{0.5em}
        \includegraphics[width=\linewidth]{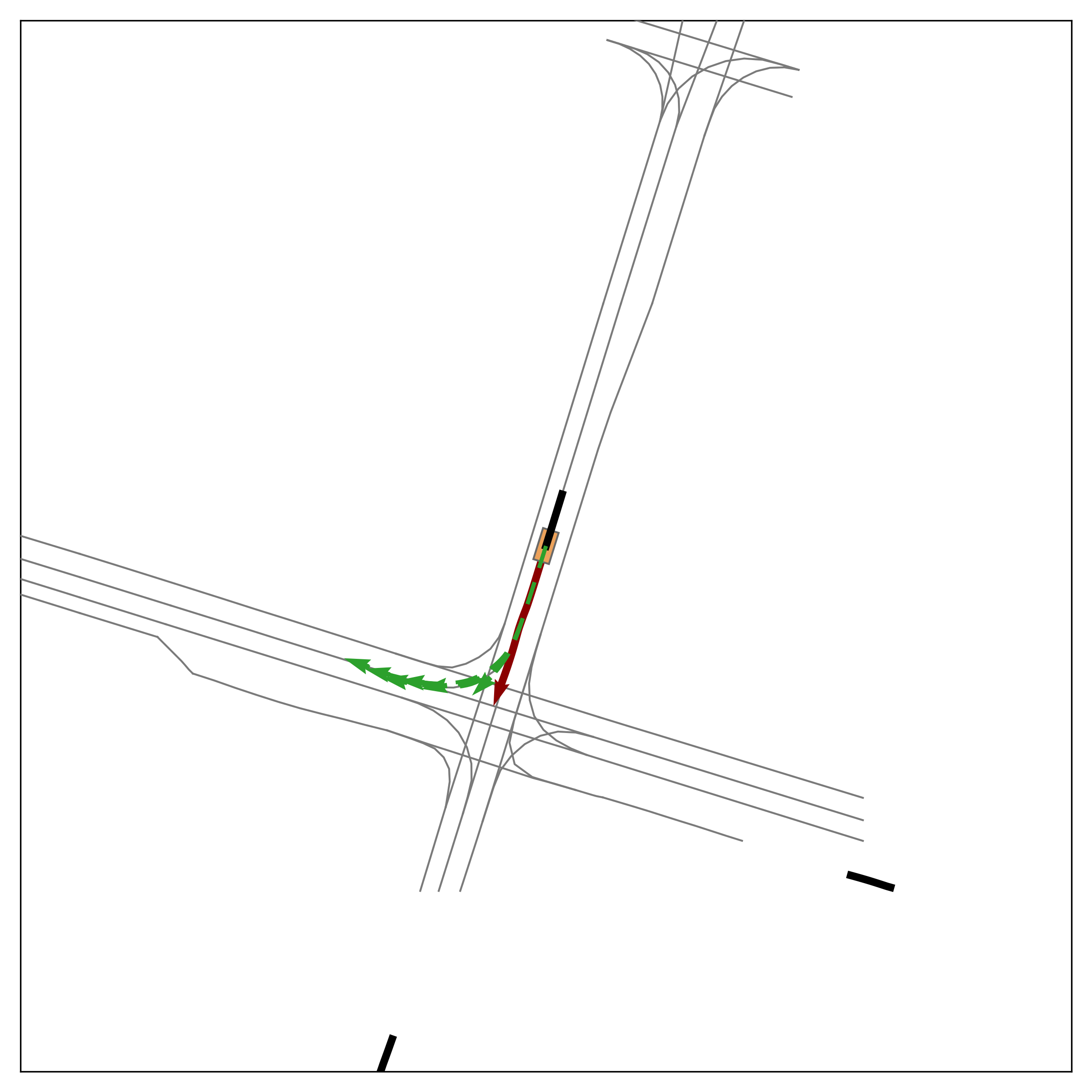}\par\vspace{0.5em}
        \includegraphics[width=\linewidth]{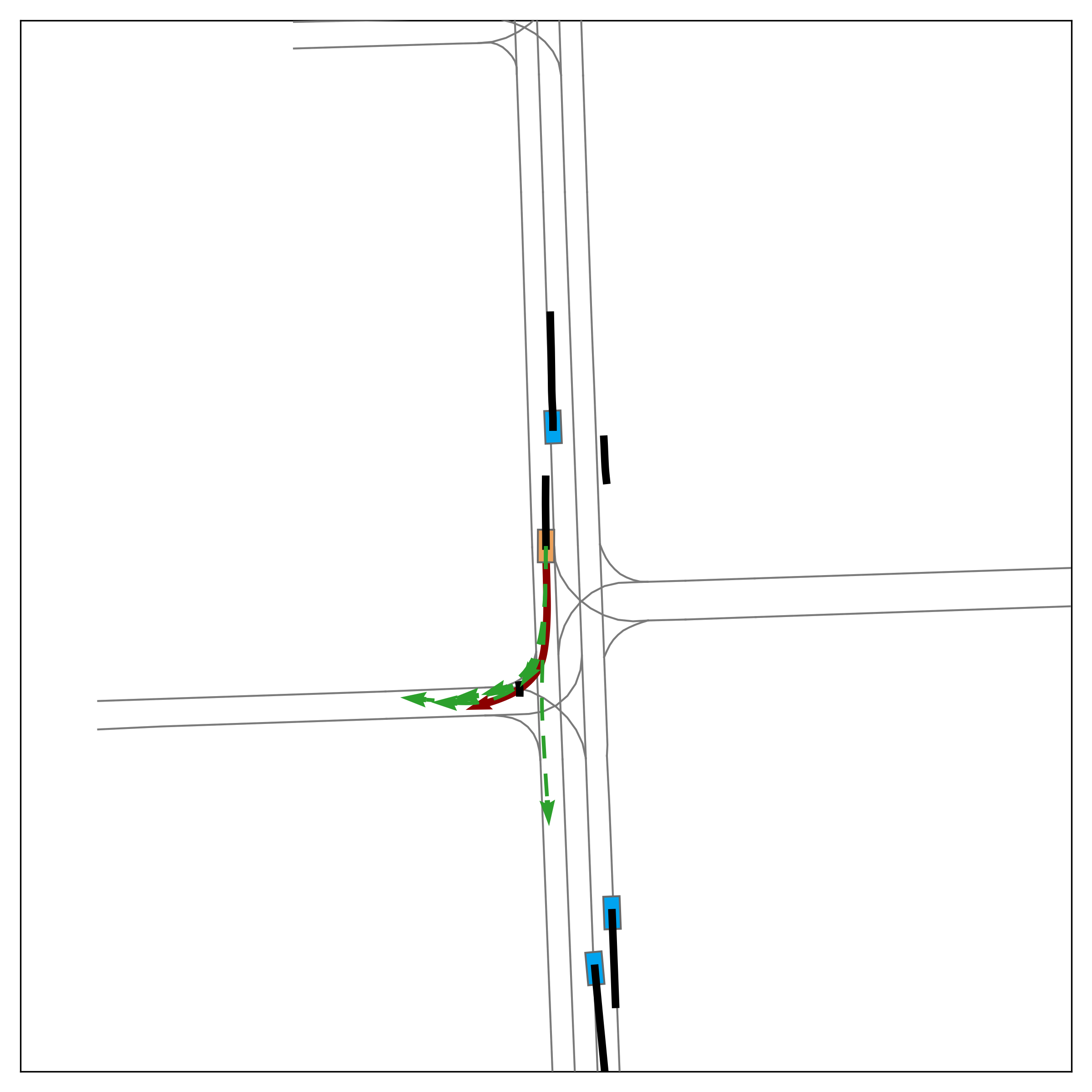}
        \caption{DeMo\_IT}
    \end{subfigure}
    \hspace{0.01\textwidth}
    \begin{subfigure}[b]{0.27\textwidth}
        \centering
        \includegraphics[width=\linewidth]{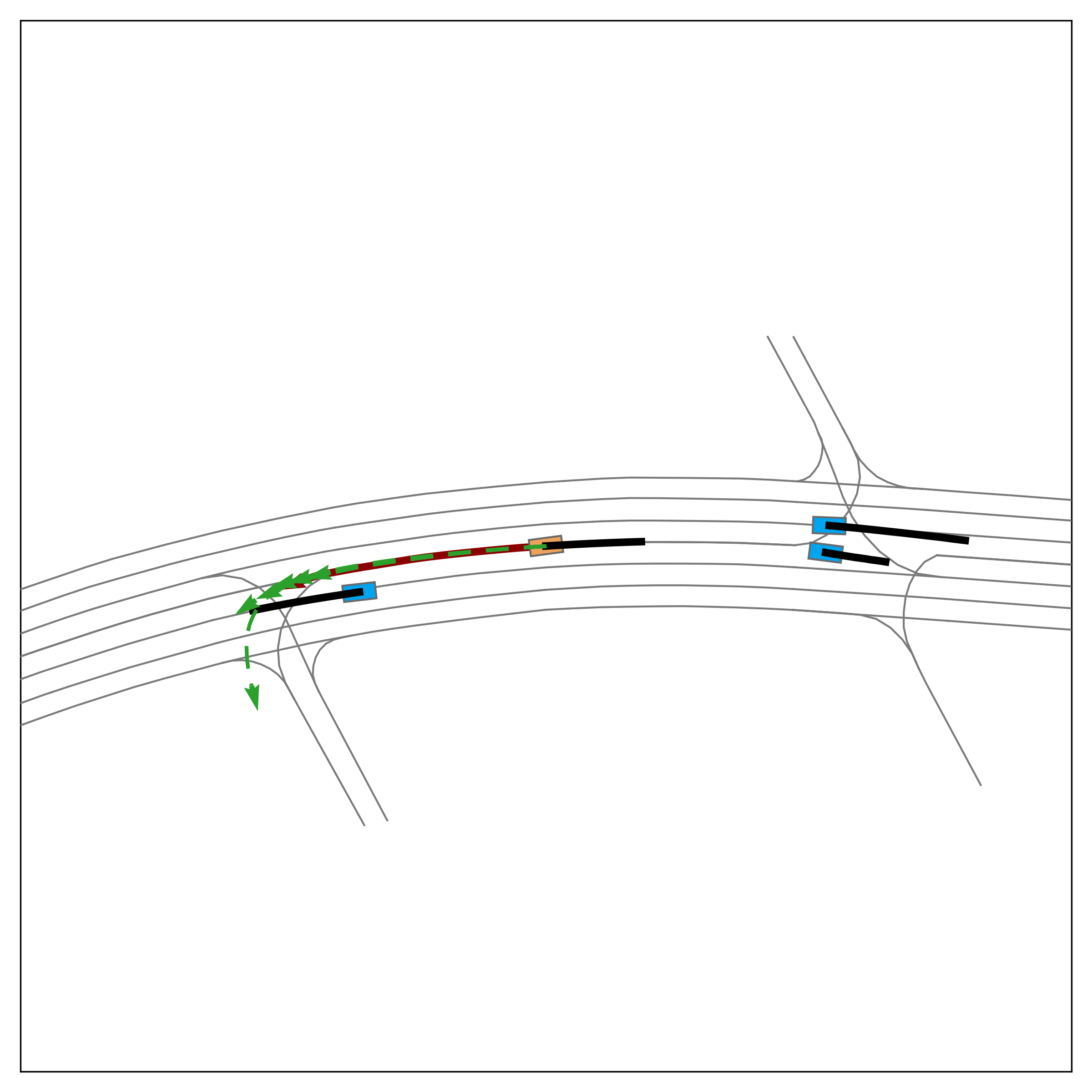}\par\vspace{0.5em}
        \includegraphics[width=\linewidth]{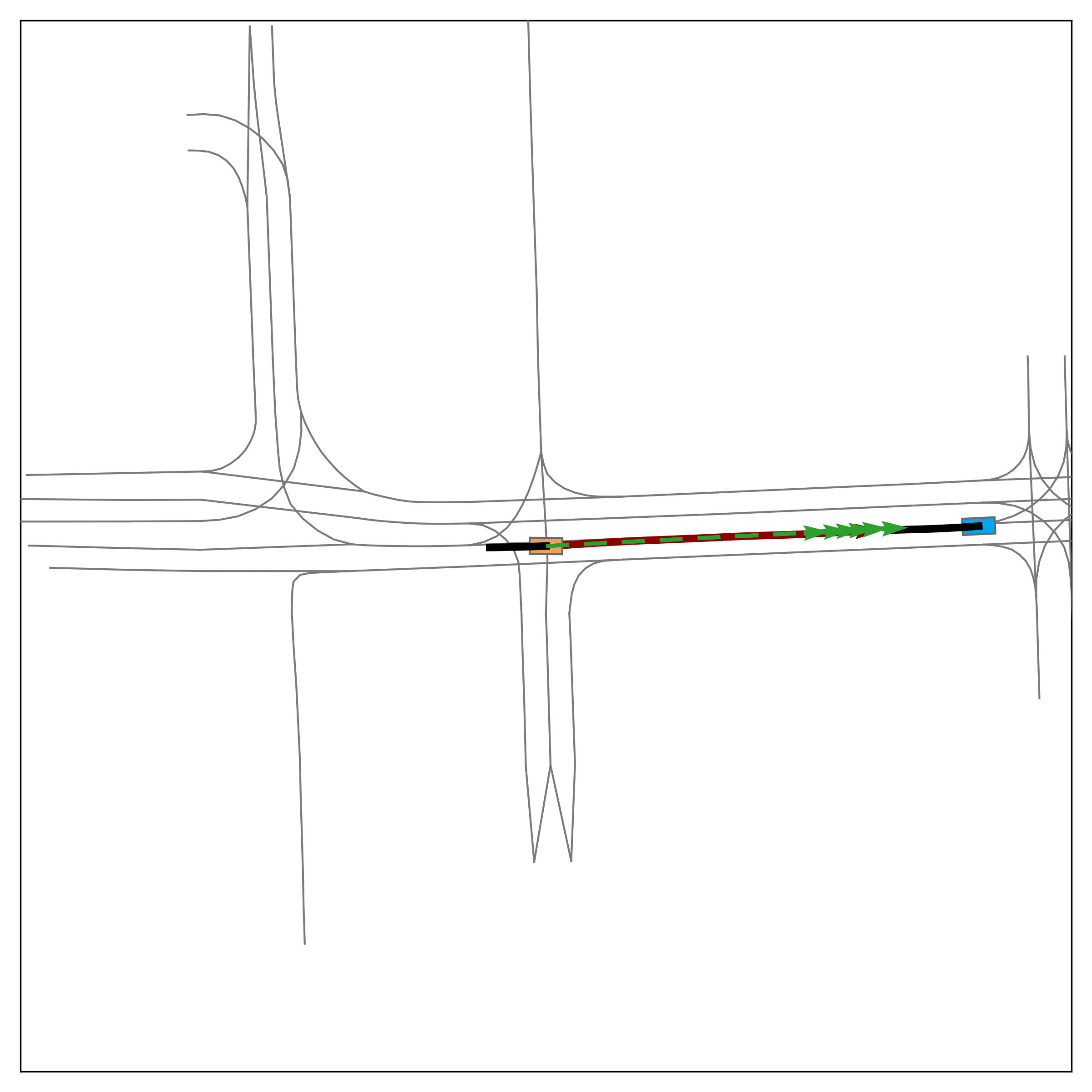}\par\vspace{0.5em}
        \includegraphics[width=\linewidth]{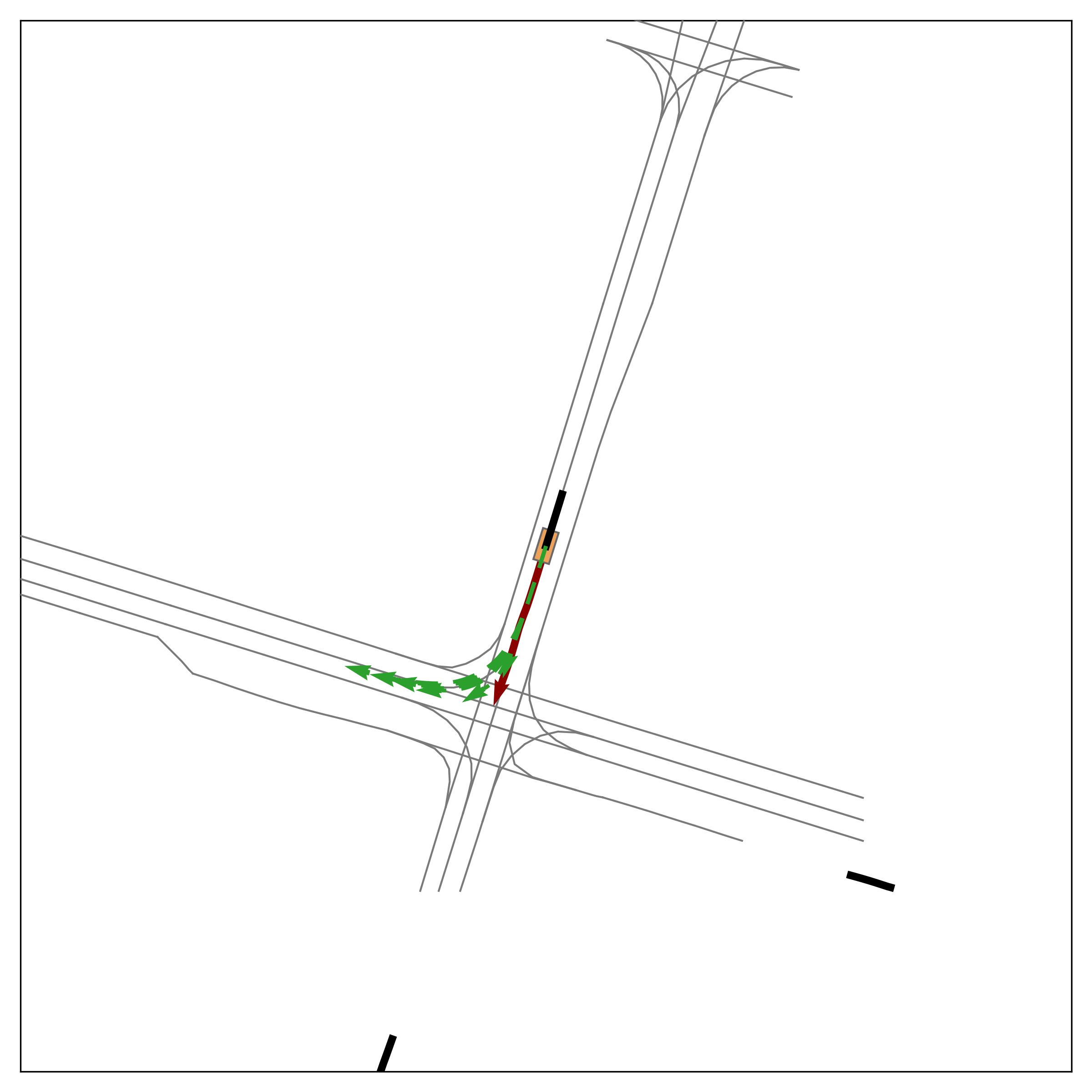}\par\vspace{0.5em}
        \includegraphics[width=\linewidth]{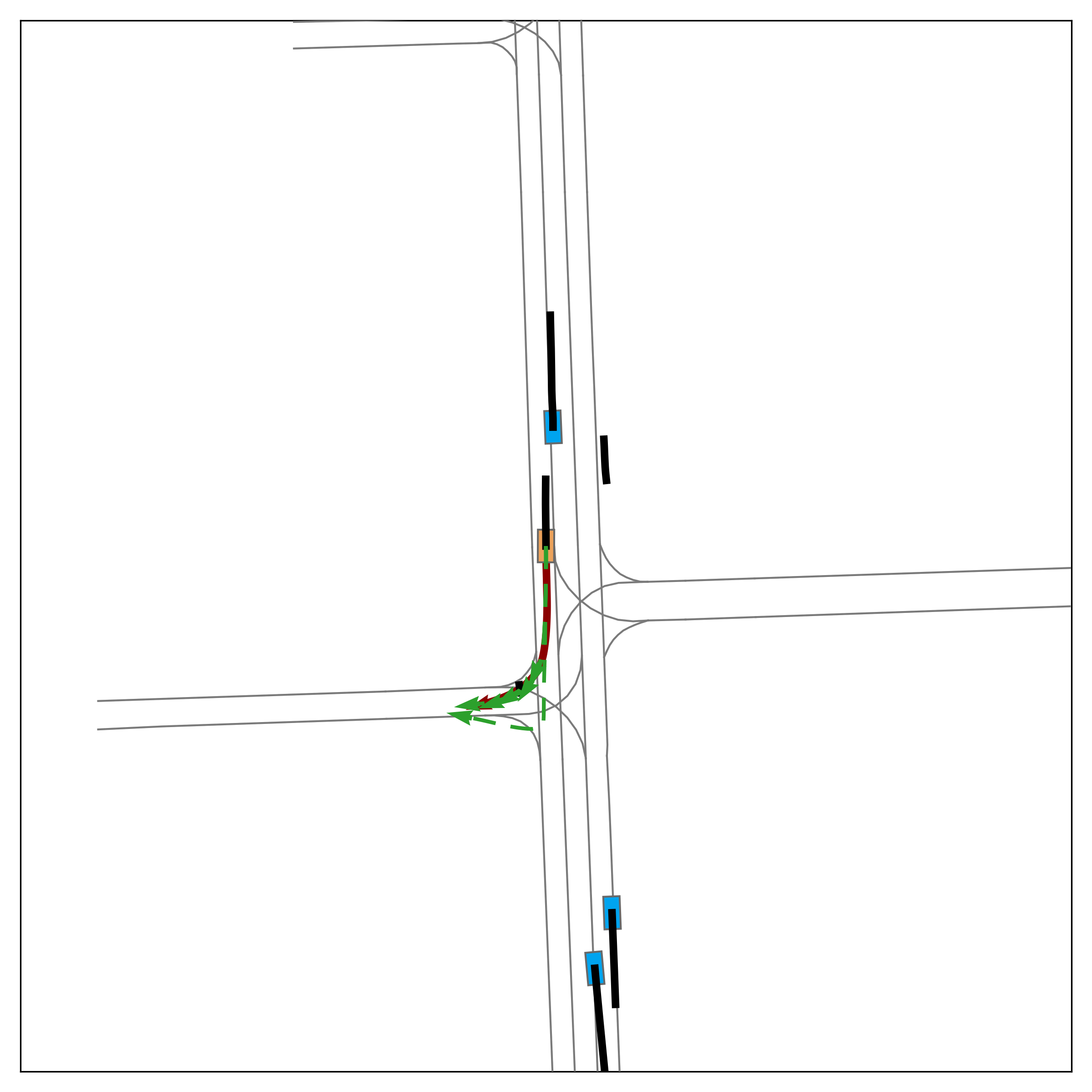}
        \caption{DeMo\_OAF}
    \end{subfigure}
    \hspace{0.01\textwidth}
    \begin{subfigure}[b]{0.27\textwidth}
        \centering
        \includegraphics[width=\linewidth]{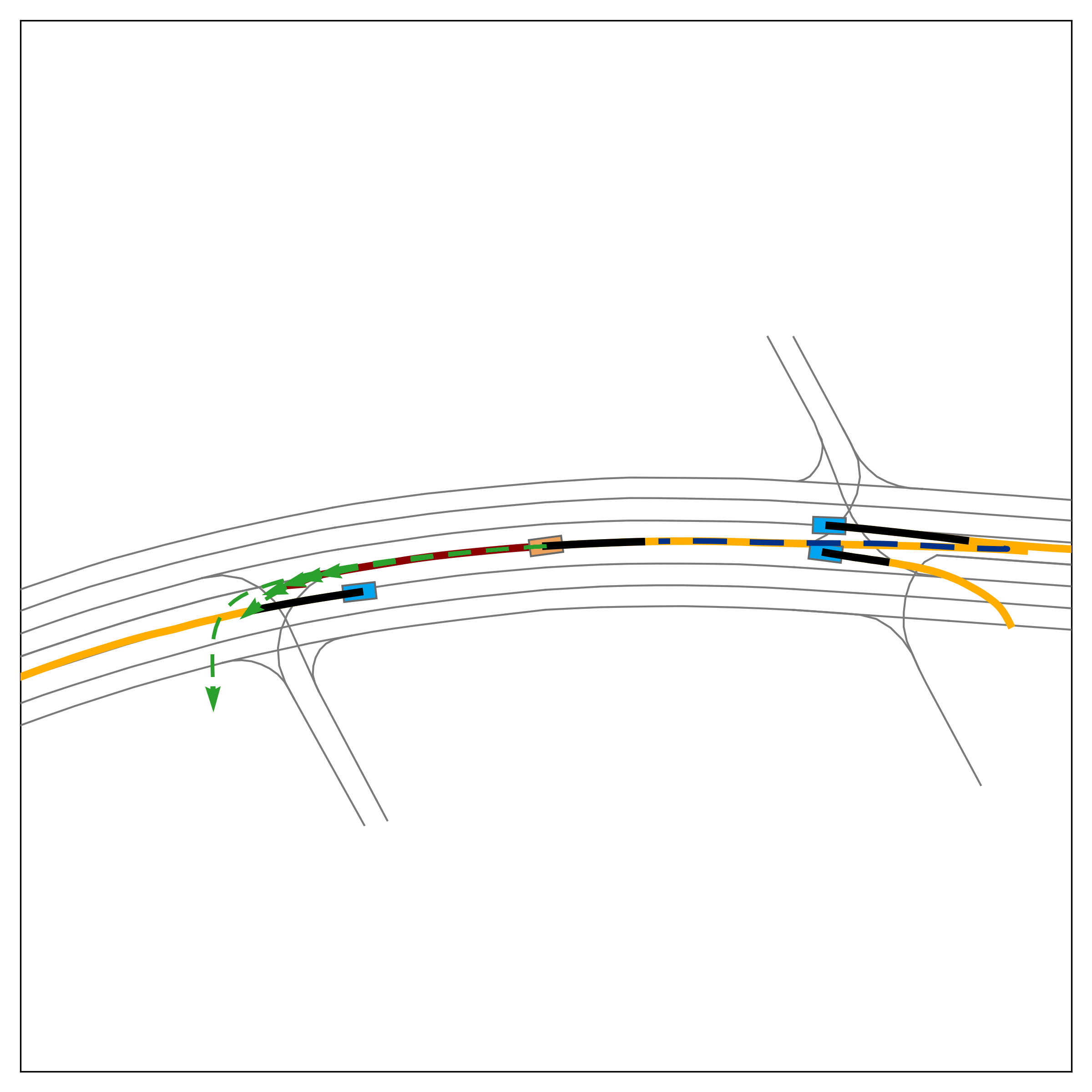}\par\vspace{0.5em}
        \includegraphics[width=\linewidth]{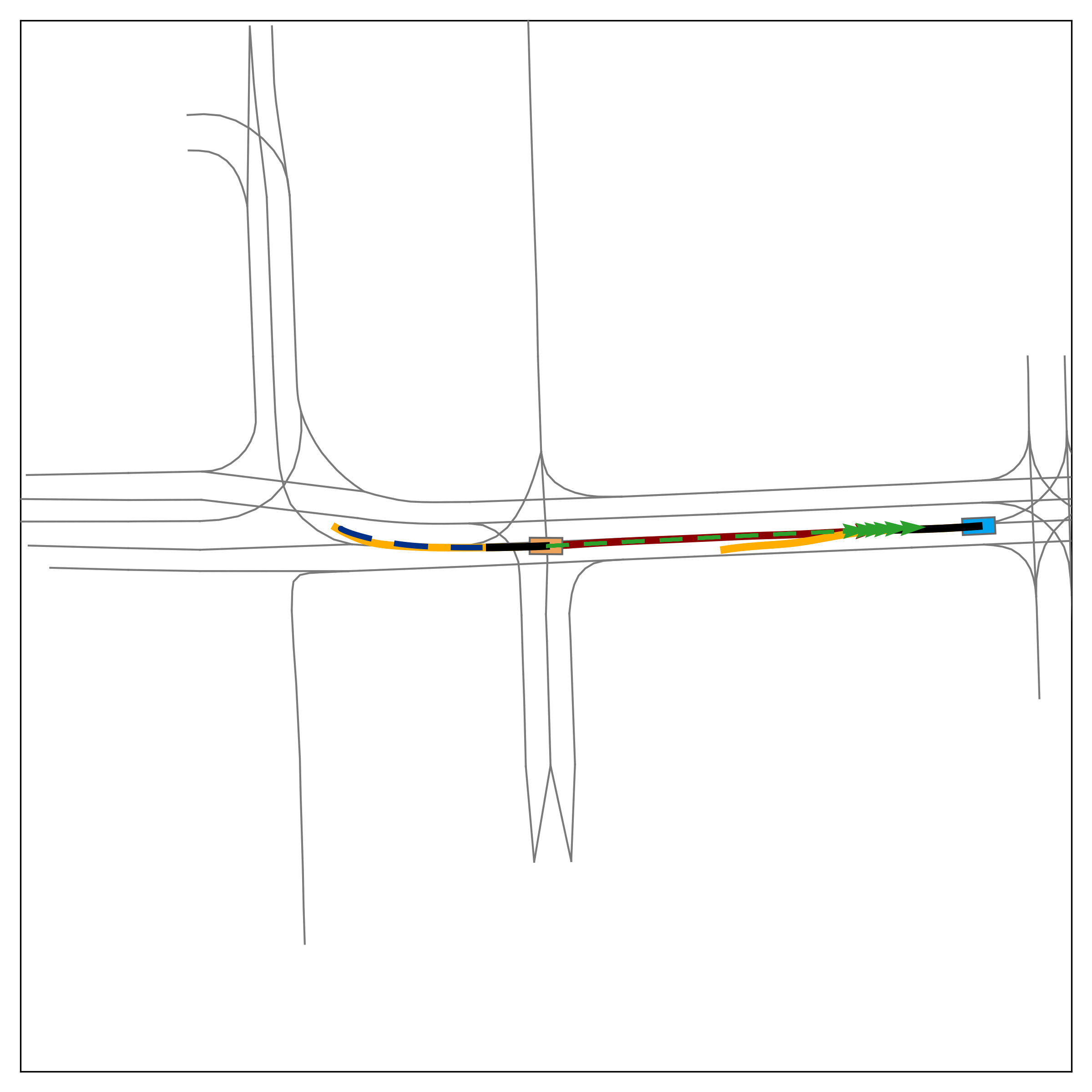}\par\vspace{0.5em}
        \includegraphics[width=\linewidth]{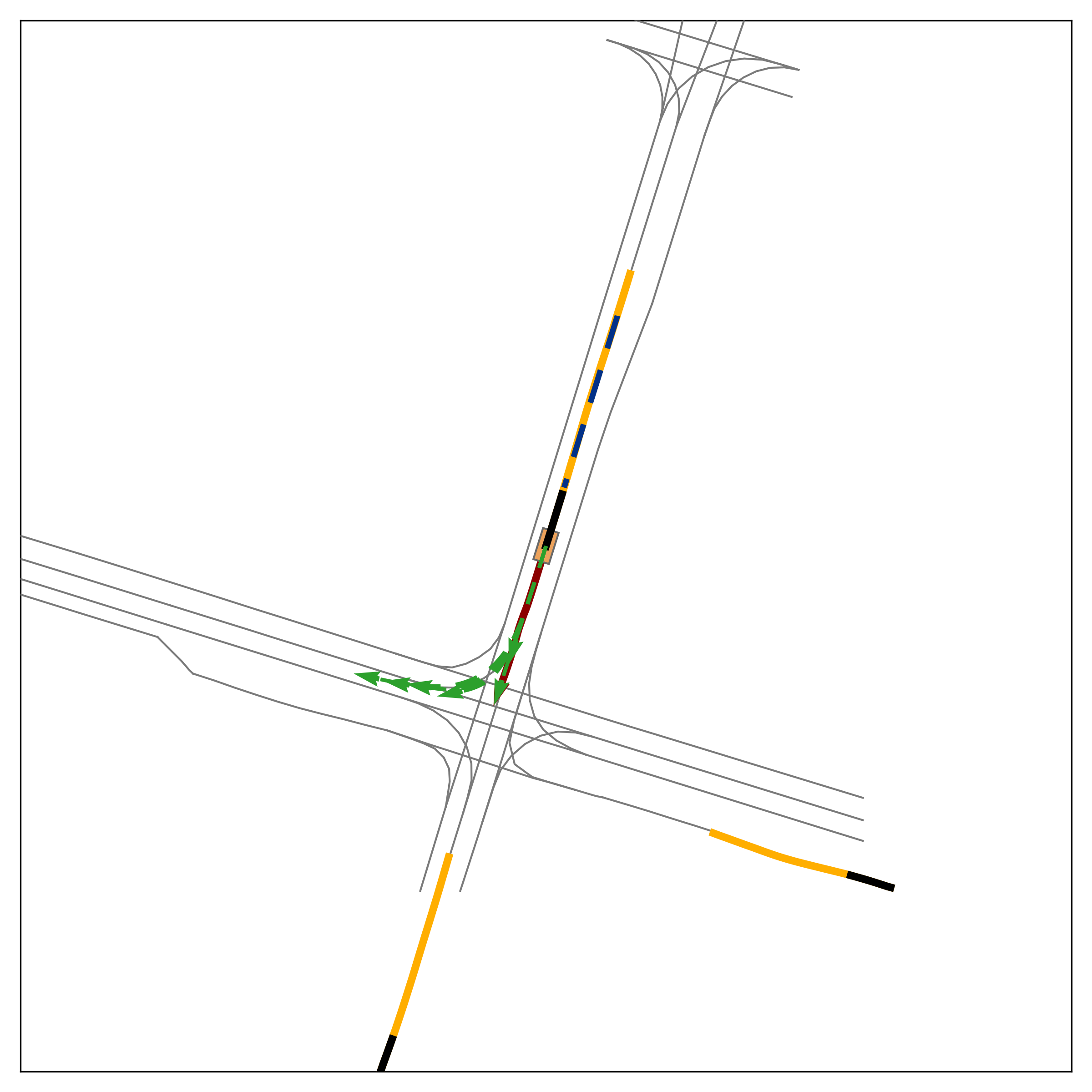}\par\vspace{0.5em}
        \includegraphics[width=\linewidth]{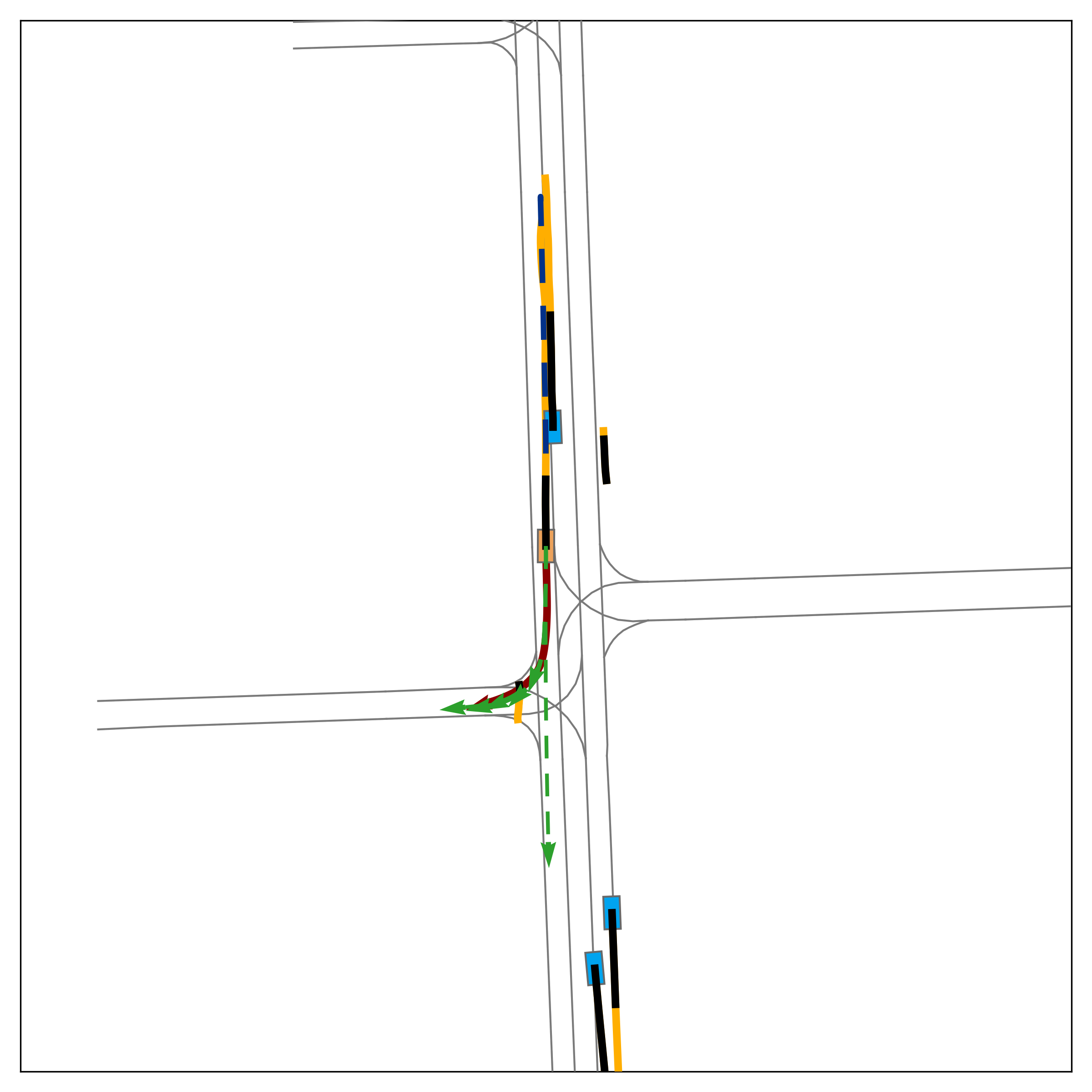}
        \caption{DeMo\_TaPD (Our full model)}
    \end{subfigure}

\caption{Qualitative comparison on the Argoverse~2 single-agent validation set under short observations (10 time steps).
Each row corresponds to one scene, and the three columns show predictions from (a) DeMo\_IT, (b) DeMo\_OAF, and (c) our DeMo\_TaPD, respectively, given the same input.
The observed history is shown as a solid black segment. The ground-truth future trajectory is shown in solid red with an arrow. Green dashed arrows denote multiple predicted trajectories.
For methods with backfilling, the reconstructed (unobserved) history is shown as a blue dashed segment, and the corresponding ground-truth unobserved history segment is shown in solid orange.}
\label{fig:visual}
\end{figure*}